\documentclass{article}

\usepackage{arxiv}
\usepackage{arxiv}
\usepackage[utf8]{inputenc} % allow utf-8 input
\usepackage{hyperref}       % hyperlinks
\usepackage{url}            % simple URL typesetting
\usepackage{booktabs}       % professional-quality tables
\usepackage{amsfonts}       % blackboard math symbols
\usepackage{nicefrac}       % compact symbols for 1/2, etc.
\usepackage{microtype}      % microtypography
\usepackage{lipsum}		% Can be removed after putting your text content
\usepackage{graphicx}
\usepackage{doi}

\usepackage{amsmath,amsfonts}
\usepackage{algorithm}
\usepackage{array}
\usepackage[caption=false,font=normalsize,labelfont=sf,textfont=sf]{subfig}
\usepackage{textcomp}
\usepackage{stfloats}
\usepackage{url}
\usepackage{verbatim}
\usepackage{graphicx}

\usepackage[acronym,nowarn,section,nogroupskip,nonumberlist]{glossaries}

\glsdisablehyper{}

\newacronym{gnn}{GNN}{Graph Neural Network}
\newacronym{ssl}{SSL}{Self-Supervised Learning}
\newacronym{fc}{FC}{Functional Connectivity}
\newacronym{fmri}{fMRI}{functional Magnetic Resonance Imaging}
\newacronym{pet}{PET}{Positron Emission Tomography}
\newacronym{roi}{ROI}{Regions of Interest}
\newacronym{vae}{VAE}{Variational Auto Encoder}
\newacronym{sce}{SCE}{Scaled Cosine Error}
\newacronym{bold}{BOLD}{Blood-Oxygen-Level-Dependent}
\newacronym{gru}{GRU}{Gated Recurrent Unit}
\newacronym{bce}{BCE}{Binary Cross-Entropy}
\newacronym{gin}{GIN}{Graph Isomorphism Network}
\newacronym{auroc}{AUROC}{Area Under the Receiver Operating Characteristics}
\newacronym{mae}{MAE}{Mean Absolute Error}
\newacronym{gcl}{GCL}{Graph Contrastive Learning}
\newacronym{mse}{MSE}{Mean Squared Error}
\newacronym{sce2}{SCE}{Softmax Cross-Entropy}
\newacronym{mlp}{MLP}{Multi-Layer Perceptron}
\newacronym{gcn}{GCM}{Graph Convolutional Network}
\newacronym{jepa}{JEPA}{Joint-Embedding Predictive Architecture}
\newacronym{ema}{EMA}{Exponential Moving Average}
\newacronym{nlp}{NLP}{Natural Language Processing}
\newacronym{mlm}{MLM}{Masked Language Modeling}
\newacronym{ours}{ST-JEMA}{Spatio-Temporal Joint Embedding Masked Autoencoder}

\usepackage{amsmath, amssymb, amsthm}
\usepackage{mathtools}
\usepackage{bbm}
\usepackage{dsfont}

\newcommand{\bsa}{\boldsymbol{a}}

\newcommand{\bse}{\boldsymbol{e}}

\newcommand{\bsm}{\boldsymbol{m}}

\newcommand{\bsp}{\boldsymbol{p}}

\newcommand{\bsx}{\boldsymbol{x}}

\newcommand{\bsz}{\boldsymbol{z}}

\newcommand{\bsA}{\boldsymbol{A}}

\newcommand{\bsH}{\boldsymbol{H}}

\newcommand{\bsM}{\boldsymbol{M}}

\newcommand{\bsP}{\boldsymbol{P}}

\newcommand{\bsR}{\boldsymbol{R}}

\newcommand{\bsW}{\boldsymbol{W}}
\newcommand{\bsX}{\boldsymbol{X}}
\newcommand{\bsY}{\boldsymbol{Y}}
\newcommand{\bsZ}{\boldsymbol{Z}}

\newcommand{\calE}{{\mathcal{E}}}

\newcommand{\calG}{{\mathcal{G}}}

\newcommand{\calL}{{\mathcal{L}}}

\newcommand{\calN}{{\mathcal{N}}}

\newcommand{\calT}{{\mathcal{T}}}

\newcommand{\calV}{{\mathcal{V}}}

\newcommand{\bbR}{\mathbb{R}}

\newcommand{\boldeta}{{\boldsymbol{\eta}}}
\newcommand{\btheta}{{\boldsymbol{\theta}}}

\newcommand{\bphi}{{\boldsymbol{\phi}}}

\newcommand{\bomega}{{\boldsymbol{\omega}}}

\theoremstyle{plain}% default

\theoremstyle{definition}

\theoremstyle{remark}

\def\[#1\]{\begin{equation}\begin{aligned}#1\end{aligned}\end{equation}}

\usepackage{orcidlink}

\newcommand{\cmark}{\color{green}{\ding{51}}}
\newcommand{\xmark}{\color{red}{\ding{55}}}
\usepackage{colortbl}

\usepackage{booktabs}       % professional-quality tables
\usepackage{multirow}
\usepackage{adjustbox}
\usepackage{wrapfig}

\usepackage{url}            % simple URL typesetting
\usepackage{color}          % colors
\definecolor{skyblue}{rgb}{0.85, 0.95, 1.0}
\usepackage{hyperref}       % hyperlinks

\usepackage{algpseudocode}
\usepackage{pifont}

\usepackage[numbers, sort]{natbib} 

\usepackage[nameinlink, capitalise]{cleveref}

\title{Joint-Embedding Masked Autoencoder for Self-supervised Learning of Dynamic Functional Connectivity from the Human Brain}

\author{
Jungwon Choi$^{1}$\thanks{Equal contribution. $^{\dagger}$Corresponding authors: Juho Lee (juholee@kaist.ac.kr), Byung-Hoon Kim (egyptdj@yonsei.ac.kr)}\hspace{2.5mm}\orcidlink{0009-0008-6219-0301} \quad
Hyungi Lee$^{1\ast}$\hspace{1mm}\orcidlink{0009-0008-7231-7310}\quad 
Byung-Hoon Kim$^{2\dagger}$\hspace{1mm}\orcidlink{0000-0003-2501-038X} \quad
Juho Lee$^{1,3\dagger}$\hspace{1mm}\orcidlink{0000-0002-6725-6874}\vspace{1mm}\\
$^{1}$KAIST AI, Daejeon, South Korea\\
$^{2}$Yonsei University College of Medicine, Seoul, South Korea\\
$^{3}$AITRICS, Seoul, South Korea\\
}

\date{}
\renewcommand{\shorttitle}{} %\textit{Choi et al.}}

\hypersetup{
pdftitle={Joint-Embedding Masked Autoencoder for Self-supervised Learning of Dynamic Functional Connectivity from the Human Brain},
pdfsubject={cs.LG, q-bio.NC},
pdfauthor={Jungwon Choi, Hyungi Lee, Byung-Hoon Kim, and Juho Lee},
pdfkeywords={Functional connectivity,  Dynamic brain networks,  Spatiotemporal graph,  Self-supervised learning,  Representation learning},
}

\begin{document}

%%%%%%%%%%%%%% Main Contents
\maketitle
\vspace{-2mm}
\begin{abstract}
    Graph Neural Networks (GNNs) have shown promise in learning dynamic functional connectivity for distinguishing phenotypes from human brain networks. However, collecting large-scale labeled clinical data is often resource-intensive, limiting their practical use. This motivates the use of self-supervised learning to leverage abundant unlabeled data. While generative self-supervised approaches—particularly masked autoencoders—have shown promising results in representation learning in various domains, their application to dynamic graphs for dynamic functional connectivity remains underexplored and faces challenges in capturing high-level semantic patterns evolving over time.
Here, we propose the Spatio-Temporal Joint Embedding Masked Autoencoder (ST-JEMA), which introduces a dual reconstruction objective that integrates spatial and temporal signals in dynamic brain graphs, enabling the encoder to learn temporally-aware semantic representations from unlabeled fMRI data. This design addresses key challenges in fMRI representation learning by capturing the spatio-temporal structure of dynamic functional connectivity. Unlike prior approaches that focus solely on static or spatial components, our method jointly reconstructs temporal dynamics and structural variations across time. Utilizing the large-scale UK Biobank dataset for self-supervised learning, ST-JEMA demonstrates robust representation learning performance across eight benchmark resting-state fMRI datasets, outperforming prior self-supervised methods in phenotype prediction and psychiatric diagnosis tasks—even with limited labeled data. Moreover, ST-JEMA proves resilient under temporal missing data scenarios, emphasizing the value of temporal reconstruction objectives. These findings demonstrate the effectiveness of our method for learning robust representations from fMRI data, particularly in label-scarce scenarios.
\end{abstract}


\section{Introduction}
\label{main:sec:introduction}

Recent advancements in neuroimaging research have shown a growing interest in leveraging \gls{fmri}, which captures the \gls{bold} signal to understand the intricate functionalities of the human brain~\citep{khosla20183d, yang2019functional}.
Consequently, leveraging \gls{fmri} data through \gls{fc} has gained traction in addressing various human brain-related problems using neural network models~\citep{kim2020understanding}. 
In this context, \gls{fc} refers to the temporal correlation degree between \gls{roi} of the brain, enabling the mathematical interpretation of \gls{fmri} data as graph structures, where nodes represent \glspl{roi} and edges represent the connectivity or correlation between these regions.
This enables the utilization of graph-based learning methods such as \gls{gnn} to analyze and interpret brain dynamics~\citep{gadgil2020spatio, kim2021learning}.

While integrating \gls{gnn} with \gls{fc} data has marked significant advancements in understanding brain networks, obtaining and labeling clinical \gls{fmri} data for supervised learning models poses challenges, affecting the scalability and practical application of these advanced techniques.
This challenge highlights the necessity of exploring alternative approaches capable of efficiently leveraging unlabeled data.

In response, utilizing unlabeled \gls{fmri} data through \gls{ssl} presents a promising solution to extract valuable features for downstream tasks~\citep{you2020graph, xia2022simgrace, hou2022graphmae}. 
\gls{ssl}, focused on pre-training backbone models without labels, has driven performance enhancements in diverse domains, including \gls{nlp}~\citep{devlin2018bert}, Image~\citep{chen2021empirical, he2022masked}, and Video Processing~\citep{wei2022masked}.
Among various SSL paradigms, generative approaches inspired by \gls{mlm} have shown superior performance compared to the contrastive learning methods in downstream tasks~\citep{he2022masked, wei2022masked} across a range of different tasks.

This trend has also emerged in graph domains, as demonstrated by GraphMAE~\citep{hou2022graphmae}, which applies a similar philosophy to static graphs and outperforms traditional contrastive \gls{ssl} approaches.
In the context of dynamic graphs, \citet{choi2023generative} have developed a generative SSL framework that enables foundational models to effectively capture temporal dynamics. 
When applied to \gls{fmri} datasets, this framework demonstrates enhanced performance across multiple downstream tasks, surpassing models based on static graph methodologies and contrastive learning approaches.

While recent advancements in generative \gls{ssl} have shown improved performance in learning graph data structures, existing methods tend to focus on reconstructing lower-level features~\citep{hou2022graphmae, choi2023generative}, potentially limiting their ability to grasp higher-level semantic representations. 
Focusing on low-level features hinders the ability of the model to capture generalizable representations for downstream tasks.

To tackle this limitation of current generative \gls{ssl} methods — reconstructing low-level representation in temporal graph data — we propose a novel framework called \gls{ours} for dynamic graphs such as dynamic functional connectivity of \gls{fmri} data. 
In \gls{ours}, we propose a new joint embedding reconstruction loss based on the latest \gls{ssl} frameworks known as \gls{jepa}~\citep{assran2023self}. 
The JEPA-inspired loss encourages the encoder to reconstruct latent representations rather than raw features, incorporating both spatial and temporal dimensions. 
This encourages the \gls{gnn} models to learn more advanced representations of dynamic graphs. 

We employed the extensive UK Biobank fMRI dataset, which consists of 40\,913 records, for comparing various \gls{ssl} methods without demographic characteristics (e.g. Gender), and validated our methodology by fine-tuning across eight benchmarks including non-clinical and clinical \gls{fmri} datasets.
Experiments showed that \gls{ours} effectively improves node representation learning compared to baseline \gls{ssl} methods.
This improvement was evident in its superior performance on downstream fMRI tasks.

Our key contributions are outlined as follows:

\begin{itemize}
    \item{We propose \gls{ours}, a novel framework that shifts generative SSL from low-level feature reconstruction toward high-level representation learning in dynamic graphs.}

    \item{By employing the extensive \gls{fmri} data from the UKB dataset, we achieved enhanced performance across diverse benchmarks, thereby validating the effectiveness of our approach in using large-scale, unlabeled data for backbone \gls{gnn} models.}

    \item{Our proposed method demonstrates enhanced efficacy in scenarios with limited labeled samples, such as with clinical datasets, notably in the data-efficient model development for psychiatric diagnosis classification.}
\end{itemize}

\begin{figure*}[t]
    \centering
    \includegraphics[width=\textwidth]
    {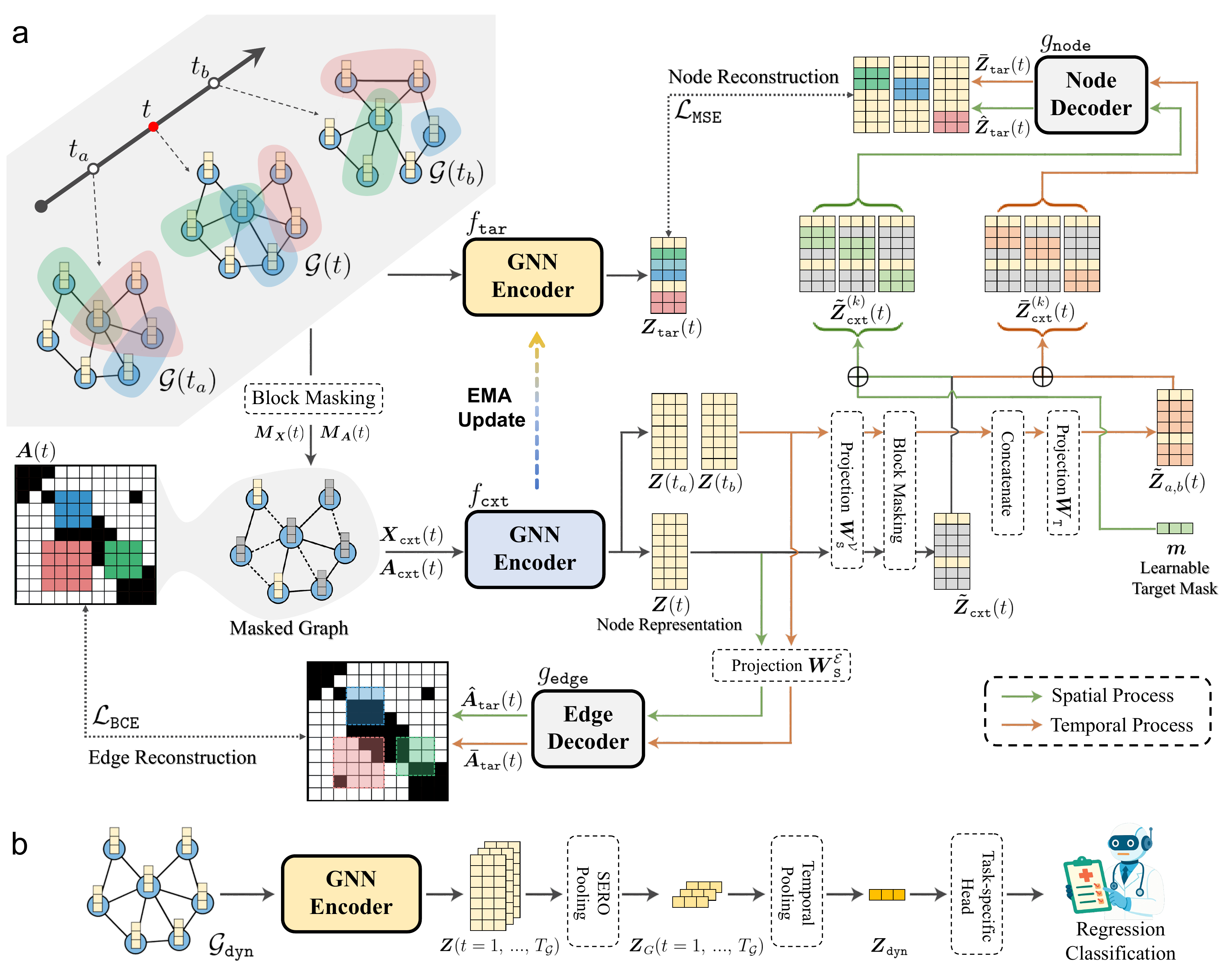}
    \caption{Overview of the Spatio-Temporal Joint Embedding Masked Autoencoder (ST-JEMA) framework.
    (a) Pre-training pipeline, where the model jointly reconstructs node representations and edge structures by masking blocks in both spatial and temporal dimensions. The context and target GNN encoders are updated via EMA. Spatial and temporal processes are indicated by green and orange dashed lines, respectively.
    (b) Fine-tuning pipeline for downstream tasks (classification or regression), where the pre-trained encoder outputs are pooled across spatial and temporal dimensions to generate task-specific representations.}
    \label{fig:overview}
\end{figure*}
\section{Related Works}
\label{main:sec:relatedworks} 

\subsection{Learning Functional Connectivity of the Brain with Graph Neural Networks }

Recent advances in \glspl{gnn} have facilitated the modeling of brain networks using \gls{fc} derived from \gls{fmri} data.
Pioneering GNN-based approaches tackle the high-dimensional nature of \gls{fmri} by modeling brain dynamics~\citep{yan2019groupinn}, employing ROI-aware architectures~\citep{li2021braingnn}, or introducing generative frameworks~\citep{kan2022fbnetgen}.

Contrary to the prevalent focus on static \gls{fc}, there have been efforts to capture the temporal dynamics within \gls{fmri} data by proposing temporal dynamic learning~\citep{wang2021modeling}, modeling temporal variants in brain networks~\citep{kim2021learning}, and combining graph structure learning with attention-based graph pooling~\citep{liu2023braintgl}, leveraging the rich temporal variation inherent to such dynamic brain networks.
These studies consider temporal variation as an important part of analyzing the brain or determining disorders from \gls{fmri} data.
In this work, we address the problem of enabling GNN encoders to learn the temporal dynamics inherent in large-scale unlabeled \gls{fmri} data by focusing on dynamic graphs. 

\subsection{Self-supervised Learning on Graph Data}

\gls{ssl} effectively extracts meaningful representations from large-scale unlabeled graph data.
In the realm of graph contrastive \gls{ssl}, Deep Graph Infomax~\citep{velivckovic2018deep} and GraphCL~\citep{you2020graph} have been pioneering efforts that leverage augmentations and multiple graph views to enrich representation learning. 
However, challenges in augmentation efficacy across domains have led to innovative works, which simplify the process by introducing perturbations to embeddings~\citep{yu2022graph} or model parameters~\citep{xia2022simgrace}.
Generative approaches in graph self-supervised learning focus on reconstructing the original graph structure or node features from encoded representations to capture inherent structural relationships. VGAE~\citep{kipf2016variational} employs variational autoencoders for adjacency matrix reconstruction enhancing link prediction performance, while 
GraphMAE~\citep{hou2022graphmae} employs a masked autoencoder approach, which shows enhanced performance in node and graph classification compared to previous contrastive methods.

Existing graph self-supervised learning methods are well-established for static graphs but remain less explored for dynamic graphs, especially in learning from dynamic \gls{fc}, with existing applications mainly in other domains such as traffic flow~\citep{gadgil2020spatio}. Recently, a generative \gls{ssl} framework designed for dynamic graphs from \gls{fmri} data, employing a masked autoencoder for temporal reconstruction, has shown promise in improving downstream task performance~\citep{choi2023generative}.

\subsection{Joint Embedding Predictive Architecture}

In the evolving landscape of \gls{ssl} in computer vision, \citet{assran2023self} introduces the \gls{jepa}, a new framework that integrates the concepts of both Joint Embedding Architecture (JEA) in contrastive learning and generative learning process of the Masked Autoencoder~\citep[MAE;][]{he2022masked}. This approach diverges from the MAE, which 
mainly focuses on pixel-level image reconstruction. Unlike MAE, which reconstructs masked patches,
\gls{jepa} aims to avoid the potential pitfall of generating representations overly fixated on low-level details. Instead, \citet{assran2023self} utilizes a predictive model trained to infer the representations of masked target patches based on unmasked context patches, enabling the encoder to formulate representations that encapsulate higher-level semantic information beyond mere pixel details. 

Refer to \cref{app:sec:addtional_relatedwork} for more detailed discussions on the related works.
\section{Preliminaries}
\label{main:sec:background}

\subsection{Problem Definition}

We aim to develop a \gls{ssl} framework for capturing temporal dynamics in dynamic graphs constructed from large-scale, unlabeled \gls{fmri} data. We pre-train a \gls{gnn} encoder $f(\cdot;\btheta)$ and decoder $g(\cdot;\bphi)$ using our \gls{ssl} strategies, and discard the decoder during downstream fine-tuning with prediction heads $h(\cdot; \bomega)$. The objective is to obtain an encoder $f(\cdot; \btheta^*)$ that captures transferable representations captured the key patterns and dynamics within the \gls{fmri} data under limited supervision.

\subsection{Dynamic Graph Construction from fMRI data}
\label{main:subsec:constructdg}

We represent fMRI-based brain activity as a dynamic graph sequence $\calG(t) = (\calV(t), \calE(t))$ over discrete time $t \in \calT$. Each node $i \in \calV(t)$ corresponds to a brain region (ROI) defined by the Schaefer atlas~\citep{schaefer2018local}, with $N$ representing the number of ROIs. Each node is associated with a feature vector $\bsx_i(t) \in \bbR^{d_v}$. The node feature is defined as:
\begin{equation}
\bsx_i(t) = \bsW[\bse_i \,\Vert\, \boldeta(t)],
\end{equation}
where $\bse_i$ is a spatial embedding and $\boldeta(t)$ is a temporal encoding from a GRU~\citep{chung2014empirical} module. 

Edges between nodes are defined using functional connectivity computed via the Pearson correlation of BOLD signals within a sliding temporal window:
\begin{equation}
R_{ij}(t) = \frac{\text{Cov}(\bar{\bsp}_i(t), \bar{\bsp}_j(t))}
{\sqrt{\smash[b]{\text{Var}(\bar{\bsp}_i(t))}}\sqrt{\smash[b]{\text{Var}(\bar{\bsp}_j(t))}}} \in \bbR^{N \times N},
\end{equation}
The adjacency matrix $\bsA(t) \in \{0,1\}^{N \times N}$ is then obtained by thresholding the top-30$\%$ of correlations. Further construction details are in Appendix~\ref{app:sec:constructdg}.

\subsection{Context and Target for Reconstruction}
\label{main:subsec:jepa}

Our method expands generative \gls{ssl} by adapting the core idea of \gls{jepa}~\citep{assran2023self}—reconstructing latent representations instead of input features—into the domain of dynamic graphs derived from \gls{fmri} data. While \gls{jepa} operates on static image patches, we introduce a multi-block masking strategy suited to spatio-temporal brain graphs. Each graph $\calG(t)$ is partially masked with $K$ different binary masks, enabling learning from diverse context-target configurations.

Specifically, we define context and target node sets at time $t$ as:
\begin{align}
\begin{gathered}
\bsX^{(k)}_{\texttt{cxt}}(t) = \{ \bsx_i(t) \mid i \in [N],  \bsM^{(k)}_{\bsX,i}(t) = \mathbf{1}_{d_v} \}, \\
\bsX^{(k)}_{\texttt{tar}}(t) = \{ \bsx_i(t) \mid i \in [N],\bsM^{(k)}_{\bsX,i}(t) = \mathbf{0}_{d_v} \}, 
\end{gathered}
\end{align}
where $\bsM^{(k)}_{\bsX}(t) \in \{0,1\}^{N \times d_v}$ is the $k$-th binary mask. 

In parallel, we apply the same masking strategy to the adjacency matrices $\bsA(t)$, where masked edges are excluded from message passing and subsequently reconstructed using the contextual node embeddings. This allows the model to jointly learn node-level and connectivity-level representations in the latent space. 

\section{Methods}
\label{main:sec:method}

\subsection{Joint Embedding Architecture}
\label{main:subsec:architecture}

To enhance \gls{gnn} encoder to capture higher-level semantic information in spatial and temporal patterns, we have introduced a new \gls{jepa}-inspired reconstruction strategy, utilizing dual encoders and decoders.
The encoders, designated as $f_{\texttt{cxt}}(\cdot; \btheta)$ for context and $f_{\texttt{tar}}(\cdot; \bar{\btheta})$ for target, encode the context and target components, respectively, with $\btheta$ representing the learnable parameters for the context encoder and $\bar{\btheta}$ for the target encoder, which is updated based on $\btheta$. 
Meanwhile, the decoders, named $g_\texttt{node}(\cdot; \bphi_\calV)$ and $g_{\texttt{edge}}(\cdot; \bphi_{\calE})$, aim to reconstruct the target nodes' representations and the adjacency matrix, respectively. Here, $\bphi_\calV$ and $\bphi_{\calE}$ denote the parameters of each decoder, respectively.

% Using separate encoders for context and target prevents representation collapse~\citep{grill2020bootstrap, assran2023self}.
Employing separate encoders for context and target components is crucial to prevent representation collapse~\citep{grill2020bootstrap} during training~\citep{assran2023self}.
Although these encoders share identical structural designs, they follow distinct parameter update rules.
Specifically, the context encoder $f_{\texttt{cxt}}$ is updated via Stochastic Gradient Descent (SGD), driven by subsequent reconstruction loss objectives, while the target encoder $f_{\texttt{tar}}$ remains frozen.

Subsequently, The parameters of target encoder, $\bar{\btheta}$, are then updated using \gls{ema} based on the evolving parameters of $f_{\texttt{cxt}}$ as follows: 
\begin{equation}
    \bar{\btheta}_{\text{new}} = \beta\bar{\btheta}_{\text{old}} + (1-\beta)\btheta_{\text{cxt}},
\end{equation}
where hyperparameter $\beta \in [0, 1]$ represents the decay factor. 
Here, $\bar{\btheta}_{\text{old}}$ and $\bar{\btheta}_{\text{new}}$ refer to the parameters of $f_{\texttt{tar}}$ before and after applying  \gls{ema} process, respectively, and $\btheta_{\text{ctx}}$ indicates the current parameters of $f_{\texttt{cxt}}$. 
Adopting distinct update rules for each encoder helps avoid simplistic and non-informative trivial representations.
This approach allows us to harness \gls{gnn} encoder to learn high-level semantic representations while maintaining nearly the same computational demand during back-propagation. 

\subsection{Block Masking Process for Masked Autoencoding}
\label{main:subsec:blockmasking}

To effectively train spatial and temporal patterns across individual time steps $t$, suitable masks need to be applied to both the node features and the adjacency matrix $\big( \bsX(t), \bsA(t) \big)$, resulting in masked versions $\big(\bsX_{\texttt{cxt}}(t), \bsA_{\texttt{cxt}}(t)\big)$ for each time step $t$.

As presented in \citet{assran2023self}, when training a model with a \gls{jepa}-based masked autoencoding objective, a sufficiently large size of the target masks, with sequential indices, is crucial in the image dataset. 
This strategy ensures that the masked targets encompass meaningful semantic information, allowing models to grasp the semantic structure by reconstructing the target data from the remaining context. 
While patch-based masking is straightforward for images, applying similar masking techniques to nodes and edges in graph data is challenging due to the absence of clear relationships between sequential indices for nodes. 
However, in the context of dynamic \gls{fc} of \gls{fmri} data, the graph construction method outlined in \citet{kim2021learning} defines a well-aligned construction between nodes with adjacent indices across time. Consequently, masking nodes with sequential indices is a natural and effective approach for the dynamic \gls{fc} of \gls{fmri} data. This method allows for the masking of nodes without sacrificing meaningful semantic information.

As mentioned earlier, generating adequately sized masks for both nodes and edges at each time step $t$ is essential to capture meaningful semantic information within the masked blocks. 
To achieve this for node features $\bsX(t)$, we generate $K$ binary mask matrices $\bsM_{\bsX}^{(k)}(t)$, where each $\bsM_{\bsX}^{(k)}(t)$ randomly and independently masks a $\lfloor\alpha_{\texttt{node}}^{(k)}(t)\cdot N\rfloor$ number of nodes with sequential indices. 
Here, mask ratio $\alpha_{\texttt{node}}^{(k)}(t)$ uniformly sampled from $(\alpha_{\text{min}}, \alpha_{\text{max}})$ where $0<\alpha_{\text{min}}<\alpha_{\text{max}}<1$ are predefined hyperparameters. This process ensures the creation of appropriately sized blocks. Refer to \cref{app:subsub:ablation_block_mask} for the experimental results regarding the ablation study on block mask ratios.

After generating $K$ distinct masks, when creating context node features for each target node feature, the memory requirements for context node features increase linearly with the growth of $K$. To remedy this issue, we employ a single global context node feature for each time $t$, outlined as follows:
\begin{align}
    \bsX_{\texttt{cxt}}(t) = \bsX(t) \odot \bigcap_{k=1}^{K} \bsM^{(k)}_{\bsX}(t),
    \label{main:eq:nodemasking}
\end{align}
where $\odot$ denotes element-wise (Hadamard) product and $\bigcap$ denotes element-wise logical AND over binary matrices.

For the adjacency matrix $\bsA (t)$, we also construct $K$ binary masks $\bsM_{\bsA}^{(k)}(t)$ for $k\in [K]$ in a similar fashion. The distinction in constructing an adjacency matrix mask primarily involves blocking a randomly chosen square-shaped submatrix. To create a mask $\bsM_{\bsA}^{(k)}(t)$ by sampling a square-shaped submatrix from the adjacency matrix, we independently and randomly select two sets of sequential indices $I_1$ and $I_2$. These sets have the same cardinality of $\lfloor\alpha_{\texttt{adj}}^{(k)}(t)\cdot N\rfloor$, where $\alpha_{\texttt{adj}}^{(k)}(t)$ is uniformly sampled from $(\alpha_{\text{min}},\alpha_{\text{max}})$.
Subsequently, we nullify the $(i,j)$th entry of $\bsM_{\bsA}^{(k)}(t)$ for all $i\in I_1$ and $j\in I_2$. Following the independent and random construction of $K$ adjacency matrix masks, akin to node features, we utilize a single global context adjacency matrix for each time $t$, described as follows:
\begin{align}
    \bsA_{\texttt{cxt}}(t) = \bsA(t) \odot \bigcap_{k=1}^{K} \bsM^{(k)}_{\bsA}(t),
    \label{main:eq:edgemasking}
\end{align}

The aforementioned masking strategies for the node features and the adjacency matrix encourage the model to learn the underlying brain connectivity patterns for each time $t$ thereby enhancing the understanding of dynamics in the spatial dimension. 

\subsection{Spatial Reconstruction Process}
\label{main:subsec:spatialobjective}

Based on the masks generated from \cref{main:subsec:blockmasking}, we introduce an objective function inspired by \gls{jepa}, referred to as $\calL_\texttt{spat}$, aiming to facilitate the learning of spatial patterns within the dynamic \gls{fc} of \gls{fmri} data using \gls{gnn} representations. To create $\calL_\texttt{spat}$, we need to reconstruct the target node representations and the target adjacency matrix based on the context node representations and the context adjacency matrix. To achieve this reconstruction, we initially build the common context node representation for each time $t$ as follows:
\begin{align}
    \bsZ(t) = f_{\texttt{cxt}} \Big( \bsX_{\texttt{cxt}}(t), \bsA_{\texttt{cxt}}(t) ; \btheta \Big). 
\end{align}

Here, we can employ any \gls{gnn} models capable of accepting node features and adjacency matrices as input for our $f_{\texttt{cxt}}$. 
To reconstruct the target node representations using only the context node representations, we reapply masking to the context node representations at the indices corresponding to the masked target node representations. 
This results in generating masked context node representations $\Tilde{\bsZ}_\texttt{cxt}(t)$ computed as follows:
\begin{align}
\begin{gathered}
    \Tilde{\bsZ}(t) = \bsW_{\texttt{S}}^{\calV} \bsZ(t), \\
    \Tilde{\bsZ}_\texttt{cxt}(t) = \Tilde{\bsZ}(t) \odot \bigcap_{k=1}^{K} \bsM_{\bsX}^{(k)}(t),
\end{gathered}
\end{align}
where $\bsW_{\texttt{S}}^{\calV} \in \bbR^{d_{\texttt{dec}} \times d_{\texttt{enc}}}$ projects encoder outputs to decoder space.
Using the masked context node representations $\Tilde{\bsZ}(t)$ and any node mask $\bsM_{\bsX}^{(k)}(t)$, we derive the target-specific context node representations $\Tilde{\bsZ}_\texttt{cxt}^{(k)}(t)$ as follows:

\begin{align}
\begin{gathered}
    \Tilde{\bsZ}_\texttt{cxt}^{(k)}(t) = \Tilde{\bsZ}_\texttt{cxt}(t) + \bsm \cdot \Big( \mathds{1}_{\bsX} -\bsM_{\bsX}^{(k)}(t) \Big),
    \label{main:eq:maskedcontext}
\end{gathered}
\end{align}
where $\mathds{1}_{\bsX}$ represents a matrix filled with ones having the same shape as $\bsM^{(k)}_{\bsX}(t)$, and $\bsm \in \bbR^{d_\texttt{dec}}$ denotes a learnable mask vector conditioning the target block we wish to reconstruct following \citet{assran2023self}. This vector is crucial for conditioning the specific target block among $K$ mask blocks from the single global context node representation $\Tilde{\bsZ}_\texttt{cxt}(t)$.

In summary, these target-specific context node representations $\Tilde{\bsZ}_\texttt{cxt}^{(k)}(t)$ are set to zero for all target node representations belonging to $K-1$ masks, except for those associated with the $k$-th mask. For the latter, corresponding representations are set to the values from the learnable mask vector $\bsm$. 
Then finally with this target-specific context node representation $\Tilde{\bsZ}_\texttt{cxt}^{(k)}(t)$, we recover the target node representation $\Hat{\bsZ}_{\texttt{tar}}^{(k)}(t)$ for the $k$-th mask as follows:
\begin{align}
    \Hat{\bsZ}_{\texttt{tar}}^{(k)}(t) = g_{\texttt{node}} \bigg( \Tilde{\bsZ}_\texttt{cxt}^{(k)}(t); \bphi_\calV \bigg).
\end{align}

For the architecture of $g_{\texttt{node}}$, we opt for the MLP-Mixer~\citep{tolstikhin2021mlp} architecture. This choice is different from the usual selection of a Multi-Layer Perceptron (MLP) as the decoder for the graph \gls{ssl}. In contrast to a simple MLP structure, the MLP-Mixer structure facilitates interactions among all nodes, enabling the reconstruction of the target node's representation by fully considering the representations of context nodes.

Our target node representations for the $k$-th mask $\bsZ_{\texttt{tar}}^{(k)}(t)$ is computed with unmasked $\bsX(t)$ and $\bsA(t)$:
\begin{align}
\begin{gathered}
    \bsZ_{\texttt{tar}}(t) = f_{\texttt{tar}} \bigg(\bsX(t), \bsA(t); \bar{\btheta} \bigg), \\
    \bsZ_\texttt{tar}^{(k)}(t) =  \bsZ_\texttt{tar}(t) \odot \Big( \mathds{1}_{\bsX} -\bsM_{\bsX}^{(k)}(t) \Big).
\end{gathered}
\end{align}
Then finally our node reconstruction objective 
$\calL_{\texttt{node-spat}}^{(t)}$ at time $t$ is computed as follows:
\begin{align}
    \calL_{\texttt{node-spat}}^{(t)} = 
    \frac{1}{K} \sum_{k\in [K]} \calL_{\texttt{MSE}}\Big(\Hat{\bsZ}_{\texttt{tar}}^{(k)}(t), \bsZ_{\texttt{tar}}^{(k)}(t)\Big),
\end{align}
where $\calL_{\texttt{MSE}}$ denotes the \gls{mse} loss. In contrast to prior graph-based MAE methods~\citep{tan2022mgae, hou2022graphmae}, where the entire node representations are recovered by the decoder module, our approach only focuses on recovering the target node representations. 
This encourages the model to focus on semantic information embedded within the masked target nodes.

To recover the adjacency matrix at time $t$, we used the common context node representation $\bsZ(t)$. Based on the reconstruction approach in \citet{kipf2016variational}, our reconstructed target adjacency matrix $\hat{\bsA}_\texttt{tar}^{(k)}(t)$ for the $k$-th mask $\bsM^{(k)}_{\bsA}(t)$ at time $t$ computed as follows:
\begin{align}
\begin{gathered}
    \bsH(t) = g_{\texttt{edge}} \Big( \bsW_{\texttt{S}}^\calE \bsZ(t) ; \bphi_\calE \Big),\\
     \hat{\bsA}_\texttt{tar}^{(k)}(t) = \texttt{sig} \Big( \bsH(t)\bsH(t)^\top \Big) \odot \Big(\mathds{1}_{\bsA} -\bsM_{\bsA}^{(k)}(t) \Big),
\end{gathered}
\end{align}
where $\mathds{1}_{\bsA}$ represents a matrix filled with ones having the same shape as $\bsM^{(k)}_{\bsA}(t)$, and $\texttt{sig}$ is an element-wise sigmoid function.
We also apply a learnable projection matrix $\bsW_\texttt{S}^\calE \in \bbR^{d_\texttt{dec} \times d_\texttt{enc}}$ to adjust dimensions from the encoder to the decoder.
And our target adjacency submatrix $\bsA_\texttt{tar}^{(k)}(t)$ for the $k$-th mask at time $t$ is simply computed as:
\begin{align}
    \bsA_\texttt{tar}^{(k)}(t) = \bsA(t) \odot \Big( \mathds{1}_{\bsA} -\bsM_{\bsA}^{(k)}(t) \Big).
\end{align}
Then our adjacency matrix reconstruction loss $\calL_{\texttt{adj-spat}}^{(t)}$ at time $t$ is computed as follows:
\begin{align}
    \calL_{\texttt{adj-spat}}^{(t)} = 
    \frac{1}{K} \sum_{k\in[K]} \calL_{\texttt{BCE}} \Big( \hat{\bsA}_\texttt{tar}^{(k)}(t), \bsA_\texttt{tar}^{(k)}(t) \Big),
\end{align}
where $\calL_{\texttt{BCE}}$ denote the \gls{bce} loss function.

In summary, our overall loss function $\calL_{\texttt{spat}}$ to capture and learn the spatial patterns is formulated as:
\begin{align}
    \calL_{\texttt{spat}} = \sum_{t\in \calT} \bigg(\lambda_{\texttt{node}} \calL_{\texttt{node-spat}}^{(t)}+\lambda_{\texttt{adj}} \calL_{\texttt{adj-spat}}^{(t)} \bigg),
\end{align}
where $\lambda_{\texttt{node}}$ and $\lambda_{\texttt{adj}}$ serve as hyperparameters regulating the scales of the two individual loss terms.

\subsection{Temporal Reconstruction Process}
\label{main:subsec:temporalobjective}

To enhance the understanding of temporal dynamics within the dynamic \gls{fc} of \gls{fmri} data, 
we also construct a \gls{jepa}-based $\calL_{\texttt{temp}}$. This temporal reconstruction strategy is designed to predict node representations and adjacency matrices at a specific time $t$ by utilizing node representation from two random points in time, $t_a$ and $t_b$, that can encompass $t$ within their interval.
For the selection of $t_a$ and $t_b$, we adopt a uniform sampling strategy following \citet{choi2023generative}, a straightforward method for choosing two temporal points. 
An important consideration when sampling $t_a$ and $t_b$ is the use of a temporal windowing process with a length of $\Gamma$ and a stride of $S$ when constructing a series of \gls{fc} matrices from the \gls{bold} signal, as discussed in \cref{app:sec:constructdg}. 
If $t_a$ and $t_b$ are sampled too closely, there may be an overlap between the windows of $t_a$ and $t_b$, making the reconstruction task trivial. 
To ensure the challenge and effectiveness of the temporal reconstruction task, we take care to ensure that the sampled time points $t_a$ and $t_b$ do not originate from overlapping windows as follows:
\begin{align}
    t_a \overset{\text{i.i.d.}}{\sim} U_{\text{int}}([0,t-d_{\text{min}}]),\quad t_b \overset{\text{i.i.d.}}{\sim}U_{\text{int}}([t+d_{\text{min}},T_\calG]),
\end{align}
where $t$ is a target timestep, $U_{\text{int}}(I)$ is a uniform distribution for all integer values in interval $I$ and $d_{\text{min}} = \frac{1}{2}(\frac{\Gamma}{S}+1)$.

Then we first reconstruct the target node representations $\bar{\bsZ}_{\texttt{tar}}^{(k)}(t)$ for the $k$-th mask at time $t$ utilizing $\bsZ(t_a)$ and $\bsZ(t_b)$ which are the context node representation at time $t_a$ and time $t_b$, respectively. To aggregate $\bsZ(t_a)$ and $\bsZ(t_b)$ into single node representation $\Tilde{\bsZ}_{a,b}(t)$, we compute the $\Tilde{\bsZ}_{a,b}(t)$ as follows:
\begin{align}
    \Tilde{\bsZ}_{a,b}(t)=\bsW_{\texttt{T}} \Big[ \Tilde{\bsZ}(t_a)\,\Vert\,\Tilde{\bsZ}(t_b) \Big],
\end{align}
where,  $\bsW_{\texttt{T}} \in \bbR^{d_\texttt{dec} \times 2d_\texttt{enc}}$ denotes a learnable projection matrix. 
Instead of employing a learnable mask vector to fill the masked target node representations at $\Tilde{\bsZ}(t)$ 
, as described in \cref{main:eq:maskedcontext}, we utilize node representations from $t_a$ and $t_b$ to construct the target-specific context node representations $\bar{\bsZ}_\texttt{cxt}^{(k)}(t)$ for the $k$-th mask at time $t$ as follows:
\begin{align}
    \bar{\bsZ}_\texttt{cxt}^{(k)}(t)=\Tilde{\bsZ}_\texttt{cxt}(t) + \Tilde{\bsZ}_{a,b}(t) \odot  \Big( \mathds{1}_{\bsX} -\bsM_{\bsX}^{(k)}(t) \Big).
\end{align}
Employing $\bsZ_{a,b}(t)$ to populate the target node representations helps the model capture temporal dynamics more effectively, enhancing its understanding of temporal variations.
The recovered target node representations are obtained using the same decoder $g_{\texttt{node}}$ as in the spatial task:
\begin{align}
    \bar{\bsZ}_{\texttt{tar}}^{(k)}(t) = g_{\texttt{node}} \bigg( \bar{\bsZ}_\texttt{cxt}^{(k)}(t); \bphi_\calV \bigg).
\end{align}
Our temporal node reconstruction loss 
$\calL_{\texttt{node-temp}}^{(t)}$ at time $t$ is computed as:
\begin{align}
    \calL_{\texttt{node-temp}}^{(t)} = 
    \frac{1}{K} \sum_{k\in [K]} \calL_{\texttt{MSE}} \Big( \bar{\bsZ}_{\texttt{tar}}^{(k)}(t), \bsZ_{\texttt{tar}}^{(k)}(t) \Big).
\end{align}

For the adjacency matrix reconstruction task, we use $\bsZ(t_a)$ and $\bsZ(t_b)$ in different ways from the spatial reconstruction. We first construct each $\bsH(t_a)$ and $\bsH(t_b)$ using $\bsZ(t_a)$ and $\bsZ(t_b)$, respectively. Then compute the reconstructed adjacency matrix for time $t$ as:
\begin{align}
    \bar{\bsA}_{\texttt{tar}}(t) =\frac{1}{2} \bigg( \texttt{sig}(\Tilde{\bsA}_{ab}) + \texttt{sig}(\Tilde{\bsA}_{ba}) \bigg),
\end{align}
where $\Tilde{\bsA}_{ab} = \bsH(t_a)\bsH(t_b)^\top$ and $\Tilde{\bsA}_{ba} = \bsH(t_b)\bsH(t_a)^\top$. Utilizing matrix multiplication between $\bsH(t_a)$ and $\bsH(t_b)$ to reconstruct the adjacency matrix also aids our model in effectively learning complex temporal dynamics within dynamic graph data~\citep{choi2023generative}.

Our reconstructed target adjacency submatrix $\bar{\bsA}_{\texttt{tar}}^{(k)}(t)$ for the $k$-th mask at time $t$ formulated as:
\begin{align}
    \bar{\bsA}_\texttt{tar}^{(k)}(t) = \bar{\bsA}_{\texttt{tar}}(t) \odot \Big( \mathds{1}_{\bsA} -\bsM_{\bsA}^{(k)}(t) \Big).
\end{align}
Therefore our temporal adjacency matrix reconstruction loss $\calL_{\texttt{adj-temp}}^{(t)}$ can be computed as follows:
\begin{align}
    \calL_{\texttt{adj-temp}}^{(t)} = 
    \frac{1}{K} \sum_{k\in [K]} \calL_{\texttt{BCE}} \Big(\bar{\bsA}_\texttt{tar}^{(k)}(t), \bsA_\texttt{tar}^{(k)}(t) \Big),
\end{align}

In summary, our overall loss function $\calL_{\texttt{temp}}$ to understand the temporal patterns is formulated as:
\begin{align}
     \calL_{\texttt{temp}} = \sum_{t\in \calT} 
     \bigg(\lambda_{\texttt{node}} \calL_{\texttt{node-temp}}^{(t)} 
     + \lambda_{\texttt{adj}} \calL_{\texttt{adj-temp}}^{(t)} \bigg).
\end{align}
Here, we use same $\lambda_{\texttt{node}}$ and $\lambda_{\texttt{adj}}$ within $\calL_{\texttt{spat}}$. 

\subsection{Summary of the Overall ST-JEMA Pipeline}
\label{main:subsec:overallpipeline}

During each training step, we calculate both the spatial reconstruction loss, $\calL_\texttt{spat}$, and the temporal reconstruction loss, $\calL_\texttt{temp}$. 
These two distinct loss components are then aggregated to formulate the combined objective function, $\calL_\texttt{ST-JEMA}$, as follows:

\begin{equation}
\calL_\texttt{ST-JEMA} = \gamma\calL_\texttt{spat} + (1-\gamma)\calL_\texttt{temp},
\end{equation}
where the hyperparameter $\gamma\in[0,1]$ is the ratio of $\calL_\texttt{spat}$.
The overview and complete training pipeline of the \gls{ours} are described in \cref{fig:overview} and \cref{app:sec:algorithm}.
\section{Experimental Setup}
\label{main:sec:expsetup}

We utilize the UK Biobank~\citep{sudlow2015uk} as the upstream dataset, a large-scale \gls{fmri} data comprised of 40\,913 subjects, for pre-training \gls{gnn} encoder with \gls{ssl}. We evaluate the pre-trained \gls{gnn} encoder across eight downstream datasets both non-clinical and clinical rs-\gls{fmri} datasets. 
For non-clinical datasets, we evaluate gender classification and age regression tasks using ABCD ~\citep{casey2018adolescent}, and three types of HCP datasets: HCP-YA~\citep{wu20171200}, HCP-A~\citep{bookheimer2019lifespan}, and HCP-D~\citep{somerville2018lifespan}. For clinical datasets, we extended psychiatric diagnosis classification tasks to distinguish between normal control groups and different psychiatric disorder groups using 
HCP-EP~\citep{lewandowski2020neuroprogression}, ABIDE~\citep{craddock2013neuro}, ADHD200~\citep{brown2012adhd}, and COBRE~\citep{mayer2013functional}.

To show effectiveness of \gls{ours}, we compared these downstream performance against previous \gls{ssl} methods including both static graphs (DGI~\citep{velivckovic2018deep}, SimGRACE~\citep{xia2022simgrace}, GAE~\citep{kipf2016variational}, VGAE~\citep{kipf2016variational}, GraphMAE~\citep{hou2022graphmae}) and dynamic graphs (ST-DGI~\citep{opolka2019spatio}, ST-MAE~\citep{choi2023generative}).

The training process was divided into two phases: 1) pre-training, where a GNN encoder $f(;\btheta)$ and decoder $g(;\bphi)$ were learned using \gls{ssl} on the UKB dataset without phenotype labels, employing a 4-layer GIN~\citep{xu2018powerful} for encoder module and a one-layer MLP-mixer for the decoder in the \gls{ours} framework, and 2) fine-tuning, where the pre-trained encoder and a task-specific head $h(;\bomega)$, incorporating the \texttt{SERO} readout module\citep{kim2021learning}, were further trained on downstream tasks with supervised learning. 

Further details on the experimental setup are provided in the dataset specifications (\cref{app:sec:dataset}), baseline descriptions (\cref{app:sec:baseline}), and training configurations (\cref{app:sec:details}).
\section{Results}
\label{main:sec:result}

% Gender
\begin{table*}[t]
\caption{Gender classification performance across various downstream \gls{fmri} datasets. The performance is evaluated using the \gls{auroc} score, with higher scores indicating better performance.}
\begin{adjustbox}{width=\linewidth, totalheight=\textheight, keepaspectratio}
\begin{tabular}{@{}l l l c c c c c c c c c @{}}
\toprule
\multirow{2}{*}{Type of FC} & \multirow{2}{*}{Pre-training} & \multirow{2}{*}{Method Type} &  \multicolumn{8}{c}{\gls{auroc} ($\uparrow$)} & \multirow{2}{*}{Rank} \\ 
\cline{4-11}
&&& ABCD & HCP-YA &  HCP-A &  HCP-D & HCP-EP & ABIDE & ADHD200 & COBRE & \\
\cline{1-12}

\multirow{6}{*}{Static}
  & Random-Init & Supervised & 72.74 & 86.93 & 68.48 & 66.22 & 55.54 & 69.62 & 63.16 & 62.94 & 8.00 \\
\cline{2-12}
  & DGI & \multirow{2}{*}{Contrastive SSL} & 72.79 & 87.16 & 68.61 & 68.63 & 54.19 & 70.08 & 61.64 & 68.76 & 6.38 \\
  & SimGRACE &  & 72.58 & 86.70 & 67.79 & 67.12 & 57.11 & 69.01 & 61.53 & 65.69 & 8.62 \\
\cline{2-12}
  & GAE & \multirow{3}{*}{Generative SSL} & 73.00 & 87.01 & 68.43 & 68.31 & 58.15 & 70.21 & 63.60 & 66.62 & 5.88 \\
  & VGAE &  & 72.87 & 86.70 & 67.57 & 66.44 & 56.25 & 69.99 & 62.63 & 66.73 & 7.75 \\
  & GraphMAE &  & 71.72 & 86.82 & 67.43 & 67.22 & 58.83 & 68.47 & 61.54 & 68.89 & 7.62 \\
\midrule
  & Random-Init & \multirow{1}{*}{Supervised} & $\underline{83.17}$ & 92.54 & 83.18 & 73.30 & 58.41 & 72.95 & $\underline{71.25}$ & $\underline{72.54}$ & 3.12 \\
\cline{2-12}
  & ST-DGI & \multirow{1}{*}{Contrastive SSL} & 81.69 & 93.37 & $\underline{84.99}$ & 72.70 & 58.15 & $\underline{74.87}$ & 68.04 & 71.77 & 3.38 \\
\cline{2-12}
  & ST-MAE &  & 82.78 & $\underline{93.71}$ & 84.34 & $\underline{74.11}$ & $\underline{60.49}$ & 74.74 & 71.01 & 67.80 & $\underline{3.00}$ \\
\rowcolor{skyblue}
\cellcolor{white}{\multirow{-4}{*}{Dynamic}}
  & \cellcolor{white}{ST-JEMA (Ours)} & \cellcolor{white}{\multirow{-2}{*}{Generative SSL}} & $\mathbf{84.16}$ & $\mathbf{93.99}$ & $\mathbf{85.65}$ & $\mathbf{74.97}$ & $\mathbf{63.58}$ & $\mathbf{75.38}$ & $\mathbf{71.94}$ & $\mathbf{80.61}$ & $\mathbf{1.00}$ \\
  
\bottomrule
\end{tabular}
\end{adjustbox}
\label{main:tab:gender}
\end{table*}

% Age
\begin{table*}[t]
\caption{Age regression performance across various downstream \gls{fmri} datasets. Performance is evaluated using the \gls{mae} score, with lower scores indicating better performance.}
\begin{adjustbox}{width=\linewidth, totalheight=\textheight, keepaspectratio}
\begin{tabular}{@{}l l l c c c c c c c c c @{}}
\toprule
\multirow{2}{*}{Type of FC} & \multirow{2}{*}{Pre-training} & \multirow{2}{*}{Method Type} &  \multicolumn{8}{c}{\gls{mae} ($\downarrow$)} & \multirow{2}{*}{Rank} \\ 
\cline{4-11}
&&& ABCD & HCP-YA &  HCP-A &  HCP-D & HCP-EP & ABIDE & ADHD200 & COBRE & \\
\cline{1-12}

\multirow{6}{*}{Static}
  & Random-Init & Supervised & 6.45 & 3.11 & 9.38 & 2.51 & 3.33 & 4.35 & 2.04 & 9.81 & 9.38 \\
\cline{2-12}
  & DGI & \multirow{2}{*}{Contrastive SSL} & 6.45 & 3.05 & 9.29 & 2.43 & 3.21 & 4.24 & 2.02 & 9.59 & 6.12 \\
  & SimGRACE &  & 6.44 & 3.08 & 9.27 & 2.45 & 3.29 & 4.34 & 2.03 & 9.74 & 7.12 \\
\cline{2-12}
  & GAE & \multirow{3}{*}{Generative SSL} & 6.44 & 3.10 & 9.21 & 2.44 & 3.18 & 4.28 & 2.03 & 9.70 & 6.38 \\
  & VGAE &  & 6.44 & 3.13 & 9.37 & 2.45 & 3.21 & 4.25 & 2.02 & 9.40 & 6.62 \\
  & GraphMAE &  & 6.44 & 3.05 & 9.35 & 2.49 & 3.17 & 4.40 & 2.01 & 9.55 & 6.38 \\
\midrule
  & Random-Init & \multirow{1}{*}{Supervised} & 6.61 & $\underline{2.78}$ & 8.04 & 2.13 & $\underline{2.94}$ & $\underline{4.10}$ & $\underline{1.90}$ & 8.79 & 3.50 \\
\cline{2-12}
  & ST-DGI & \multirow{1}{*}{Contrastive SSL} & 6.45 & $\underline{2.78}$ & $\underline{8.01}$ & $\underline{2.10}$ & 2.96 & 4.16 & 1.91 & 9.16 & 3.50 \\
\cline{2-12}
  & ST-MAE &  & $\underline{6.43}$ & 2.82 & 8.37 & 2.13 & 3.13 & $\underline{4.10}$ & $\mathbf{1.86}$ & $\underline{8.76}$ & $\underline{2.75}$ \\
\rowcolor{skyblue}
\cellcolor{white}{\multirow{-4}{*}{Dynamic}}
  & \cellcolor{white}{ST-JEMA (Ours)} & \cellcolor{white}{\multirow{-2}{*}{Generative SSL}} & $\mathbf{6.42}$ & $\mathbf{2.73}$ & $\mathbf{7.94}$ & $\mathbf{2.09}$ & $\mathbf{2.92}$ & $\mathbf{4.08}$ & $\mathbf{1.86}$ & $\mathbf{8.60}$ & $\mathbf{1.00}$ \\
  
\bottomrule
\end{tabular}
\end{adjustbox}
\label{main:tab:age}
\end{table*}

% Psychiatric Diagnosis Classification
\begin{table*}[t]
\centering
\caption{Psychiatric diagnosis classification performance on clinical  \gls{fmri} datasets. The performance is evaluated using the \gls{auroc} score, with higher scores indicating better performance.}
\begin{small}
\setlength{\tabcolsep}{11pt}
\begin{tabular}{@{}l l l c c c c c @{}}
\toprule
\multirow{2}{*}{Type of FC} & \multirow{2}{*}{Pre-training} & \multirow{2}{*}{Method Type} &  \multicolumn{4}{c}{\gls{auroc} ($\uparrow$)} & \multirow{2}{*}{Rank} \\ 
\cline{4-7}
&&& HCP-EP & ABIDE & ADHD200 & COBRE & \\
\cline{1-8}

\multirow{6}{*}{Static}
  & Random-Init & Supervised & 67.47 & 63.04 & 55.44 & 57.86 & 8.75 \\
\cline{2-8}
  & DGI & \multirow{2}{*}{Contrastive SSL} & 67.13 & 64.30 & 56.55 & 62.09 & 6.00 \\
  & SimGRACE &  & 64.41 & 64.54 & $\underline{58.25}$ & 61.50 & 5.75 \\
\cline{2-8}
  & GAE & \multirow{3}{*}{Generative SSL} & 68.74 & 64.52 & 56.28 & $\underline{64.00}$ & 4.75 \\
  & VGAE &  & 66.79 & 64.32 & 55.74 & 61.74 & 7.25 \\
  & GraphMAE &  & 63.67 & 64.22 & 55.78 & 63.48 & 7.25 \\
\midrule
  & Random-Init & Supervised & 74.15 & 67.16 & 55.37 & 60.98 & 6.75 \\
\cline{2-8}
  & ST-DGI & \multirow{1}{*}{Contrastive SSL} & 77.78 & 68.86 & 56.46 & 61.08 & 4.75 \\
\cline{2-8}
  & ST-MAE &  & $\underline{78.66}$ & $\underline{69.73}$ & 56.77 & 62.19 & $\underline{2.75}$ \\
\rowcolor{skyblue}
\cellcolor{white}{\multirow{-4}{*}{Dynamic}}
  & \cellcolor{white}{ST-JEMA (Ours)} & \cellcolor{white}{\multirow{-2}{*}{Generative SSL}} & $\mathbf{79.22}$ & $\mathbf{71.49}$ & $\mathbf{58.89}$ & $\mathbf{70.04}$ & $\mathbf{1.00}$ \\

\bottomrule
\end{tabular}
\end{small}
\label{main:tab:diagnosis}
\end{table*}

\subsection{Downstream-task Performance}
\label{main:subsec:downstream_main}

\subsubsection{Gender Classification}

In gender classification tasks using the gender class phenotype from benchmark datasets, our proposed method, \gls{ours}, demonstrates superior AUROC performance across all benchmarks as shown in \cref{main:tab:gender}. Notably, it proved especially effective in classifying gender from clinical datasets, which typically comprise fewer samples than non-clinical datasets. We noted that employing dynamic rather than static \gls{fc} resulted in performance improvements when comparing the overall performance of static and dynamic methods. This outcome proves the significance of considering temporal variations in gender classification tasks. 
Additionally, \gls{ours} outperformed previous dynamic graph self-supervised learning methods. This enhancement in performance shows that \gls{ours} has effectively learned representations that capture the temporal dynamics in the upstream rs-\gls{fmri} data.

\subsubsection{Age Regression}

In the age regression task, \gls{ours} consistently surpassed other methods in prediction accuracy across the most of datasets. 
The evaluation metric scales vary due to different age ranges in each dataset, refer to \cref{main:tab:age}.
Interestingly, the performance of the \gls{gnn} encoder trained via ST-DGI was similar to, or in some instances worse than, the Dynamic Random-Init method, especially in clinical datasets that have fewer samples.  
Contrastive methods like ST-DGI showed unstable results, while \gls{ours} showed improved performance even in clinical data, suggesting that the representations learned through our methodology were more effective in distinguishing phenotypes.

\subsubsection{Psychiatric Diagnosis Classification}

The motivation behind employing \gls{ssl} with a large amount of unlabeled \gls{fmri} data is to improve the performance of the downstream diagnosis classification task, typically constrained by a limited number of labeled training datasets.
The \gls{gnn} encoder is expected to learn the structure of \gls{fmri} data during upstream training, enabling better psychiatric classification with minimal clinical data.
As shown in \cref{main:tab:diagnosis}, in most cases, \gls{ssl}-trained model showed better performance than those initialized randomly, proving that learning representations from unlabeled data is beneficial for accurately detecting signal variations in \gls{bold} signals for psychiatric diagnosis classification. \gls{ours} notably surpassed other methods in psychiatric diagnosis classification, especially demonstrating significantly higher \gls{auroc} in the COBRE dataset, even when tasked with distinguishing between three distinct psychiatric conditions and control groups.

We further evaluated the versatility of representations learned by \gls{ours} through linear probing, multi-task learning, and temporal missing data scenarios. These results support the robustness of our model across diverse settings.
See \cref{app:additional_experiments} for details.

\subsection{Ablation Study}

\begin{figure}[t]
        \centering
        \includegraphics[width=0.5\linewidth]{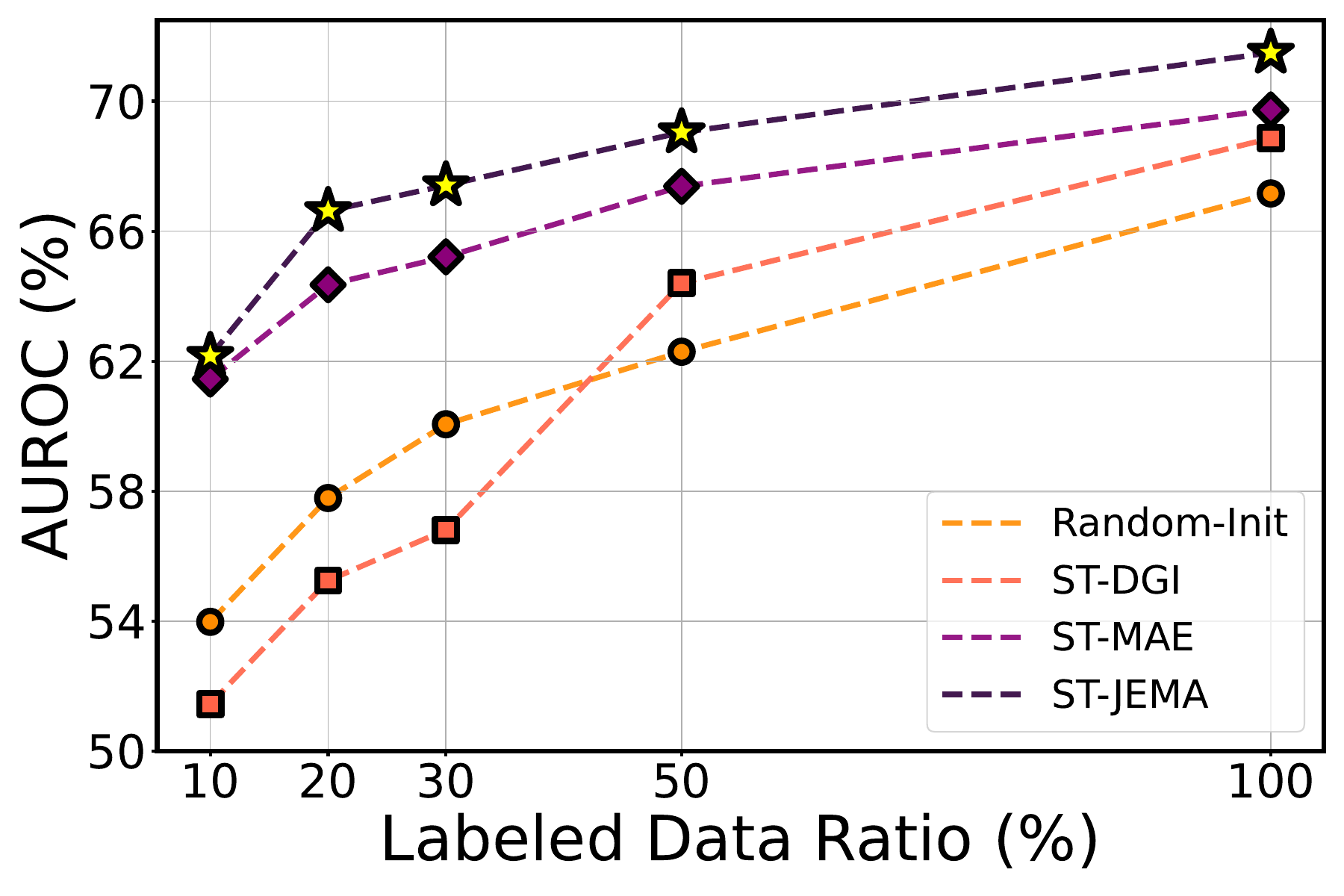}
        \caption{Ablation study on the impact of limited labeled samples for the diagnosis classification task. We fine-tuned the pre-trained \gls{gnn} model using a limited number of samples, randomly sampled from the ABIDE dataset.}
        \label{fig:limited_sample}
        \vspace{-3mm}
\end{figure}

\subsubsection{Fine-tuning Performance with Limited Samples}

Clinical \gls{fmri} datasets often suffer from limited labeled data, which hinders the training of robust models. To assess the effectiveness of our pre-trained \gls{gnn} under such constraints, we conducted experiments on $10\%$, $20\%$, $30\%$, and $50\%$ subsets of the ABIDE dataset. and compared \gls{ours} with other dynamic graph \gls{ssl} methods on psychiatric diagnosis classification.
As shown in \cref{fig:limited_sample}, \gls{ours} consistently outperformed all baselines, even with as little as $20\%$ of the training data, indicating its ability to learn semantic representations beneficial under data scarcity. Notably, ST-DGI failed to surpass Random-Init when using less than $50\%$, revealing the limitations of contrastive-based methods. These results demonstrate the effectiveness of generative \gls{ssl} for clinical neuroimaging tasks where labeled data is scarce.

\subsubsection{Ablation on Loss Function}
\label{main:subsub:ablation_loss}

\begin{table}[t]
\centering
\caption{Ablation results on reconstruction losses in $\calL_{\text{ST-JEMA}}$. Baseline (1)--(4) node, edge, spatial, and temporal loss components, respectively.}
\label{main:tab:ablation_loss}
\begin{tabular}{ c c c c c c}
\toprule

Method & Node & Edge & $\calL_{spatial}$ & $\calL_{temporal}$ & AUROC ($\uparrow$) \\
\midrule
Random Init & {\xmark} & {\xmark} & {\xmark} & {\xmark} & 67.16 \\
\midrule
Baseline (1) & {\xmark} & {\cmark} & {\cmark} & {\cmark} & 68.24 \\
Baseline (2) & {\cmark} & {\xmark} & {\cmark} & {\cmark} & 70.33 \\
Baseline (3) & {\cmark} & {\cmark} & {\xmark} & {\cmark} & 69.79 \\
Baseline (4) & {\cmark} & {\cmark} & {\cmark} & {\xmark} & 65.19 \\
\midrule
\rowcolor{skyblue}
ST-JEMA (Ours) & {\cmark} & {\cmark} & {\cmark} & {\cmark} & $\mathbf{71.49}$ \\

\bottomrule
\end{tabular}
\end{table}

In formulating the loss $\calL_{\text{ST-JEMA}}$, we consider four key components: 1) node representation reconstruction, 2) edge reconstruction, 3) spatial, and 4) temporal objectives. We evaluate their contributions through an ablation study on the psychiatric diagnosis classification task using the ABIDE dataset.
\cref{main:tab:ablation_loss} clearly shows that excluding any loss term from $\calL_{\text{ST-JEMA}}$ results in a significant drop in performance compared to \gls{ours}. Notably, the most substantial performance decline is observed when excluding the temporal loss $\calL_{\text{temp}}$, highlighting the critical importance of capturing temporal dynamics in \gls{fmri} data for accurately addressing downstream tasks. Meanwhile, the performance of baselines, except when excluding the temporal loss, consistently outperforms the Random Init baseline. 
This demonstrates the significance of pre-training on a large upstream dataset, particularly in scenarios where the downstream data is limited in terms of labeled samples, as commonly encountered in clinical \gls{fmri} datasets.

Refer to \cref{app:additional_experiments} to see further ablation experiments on the impact of pre-training data size, block mask ratio, and decoder architecture.
\section{Conclusion}
\label{main:sec:conclusion}

We introduced Spatio-Temporal Joint-Embedding Masked Autoencoder (ST-JEMA), \gls{ssl} framework tailored for \gls{fmri} data.
Leveraging the \gls{jepa}~\citep{assran2023self} framework, \gls{ours} captures both spatial and temporal dynamics in the dynamic \gls{fc} of \gls{fmri} data by jointly optimizing spatial and temporal reconstruction objectives.
Extensive experiments on both non-clinical and clinical downstream tasks demonstrate that our method outperforms existing \gls{ssl} approaches across classification and regression tasks. 
These findings highlight the effectiveness of \gls{ours} in learning robust representations from unlabeled \gls{fmri} data, particularly in data-constrained scenarios common in neuroimaging research.

\section*{Acknowledgments}
We would like to thank Sungho Keum for his valuable discussions that contributed to the development of this work.
This research was partly supported by Basic Science Research Program through the National Research Foundation of Korea (NRF) funded by the Ministry of Education (NRF-2022R1I1A1A01069589), National Research Foundation of Korea(NRF) grant funded by the Korea government(MSIT) (NRF-2021M3E5D9025030), Institute of Information \& communications Technology Planning \& Evaluation (IITP) grant funded by the Korea government(MSIT) (No.2019-0-00075, Artificial Intelligence Graduate School Program(KAIST), No.2022-0-00713, Meta-learning Applicable to Real-world Problems).

\bibliographystyle{unsrtnat}
\bibliography{references} 

%%%%%%%%%%%%%% Appendix Contents
\clearpage
\appendix 

\section{Algorithmic Flow of ST-JEMA}
\label{app:sec:algorithm}

To further facilitate understanding, we summarized the overall pipeline of our proposed Spatio-Temporal Joint Embedding Masked Autoencoder (ST-JEMA) framework for self-supervised learning on fMRI data in \cref{alg:stjema}. The algorithm describes the spatial and temporal reconstruction process of node representations and adjacency matrices employing K block masks, assuming the initial conversion of fMRI BOLD signals $\bsP$ into dynamic graphs $\calG(t)$ has already been completed. Detailed explanations of each equation and the methodological underpinnings can be found in \cref{main:sec:method}.

\begin{algorithm*}[p]
\caption{Spatio-Temporal Joint Embedding Masked Autoencoder (ST-JEMA)}
\label{alg:stjema}
\small
\begin{algorithmic}[1]
\State \textbf{Input:} Node features $ \bsX(t) $, Adjacency matrix $ \bsA(t) $ where $ t \in [T_\calG]$, the number of times for dynamic graphs $T_\calG$, minimum and maximum masking size ratio of nodes $(\alpha_{\text{min}}, \alpha_{\text{max}})$, the number of node N, window length $\Gamma$, window stride $S$, the ratio of spatial reconstruction factor $\gamma$, node reconstruction weight $\lambda_{\texttt{node}}$, and edge reconstruction weight  $\lambda_{\texttt{adj}}$.
\State \textbf{Output:} pretrained target \gls{gnn} encoder $f ( \cdot ; \bar{\theta})$
\State Initialize context encoder $f_{\texttt{cxt}}$, target encoder $f_{\texttt{tar}}$, node decoder $ g_{\texttt{node}} $ and edge decoder $ g_{\texttt{edge}} $.
\For{each step}

\State $\calL_\texttt{spat} \gets 0$ and $\calL_\texttt{temp} \gets 0$ \Comment{Initialize losses for each step}
\For{$t \in \calT$}
    \State \texttt{/* Block masking process.*/}
        \State Sample $K$ Block Masks $\bsM_{\bsX}(t), \bsM_{\bsA}(t)$ with $\alpha(t) \sim Uniform(\alpha_{\text{min}}, \alpha_{\text{max}})$. 
        \State $\bsX_{\texttt{cxt}}(t), \bsA_{\texttt{cxt}}(t) \gets \text{Block Masking}(\bsX(t), \bsA(t), \bsM_{\bsX}(t), \bsM_{\bsA}(t))$
        
        \State $\bsZ(t) \gets f_{\texttt{cxt}} \Big( \bsX_{\texttt{cxt}}(t), \bsA_{\texttt{cxt}}(t) ; \theta\Big) $ 
        \State $\bsZ_{\texttt{tar}}(t) \gets f_{\texttt{tar}} \Big( \bsX(t), \bsA(t) ; \bar{\theta}\Big) $ 
    
    \State \texttt{/* Encode context and target node features */}
        \State $\bsZ(t) \gets f_{\texttt{cxt}} \Big( \bsX_{\texttt{cxt}}(t), \bsA_{\texttt{cxt}}(t) ; \theta\Big) $ 
        \State $\bsZ_{\texttt{tar}}(t) \gets f_{\texttt{tar}} \Big( \bsX(t), \bsA(t) ; \bar{\theta}\Big) $

    \State \texttt{/* Spatial reconstruction*/}
        \State $\Tilde{\bsZ} \gets \bsW_{\texttt{S}}^{\calV} \bsZ(t)$
        \State $\Tilde{\bsZ}_\texttt{cxt}(t) \gets \Tilde{\bsZ} \odot \bigcap_{k=1}^{K} \bsM^{(k)}_{\bsX}(t)$
        \State $\bsH(t) \gets g_{\texttt{edge}} \Big( \bsW_{\texttt{S}}^{\calE}\bsZ(t) ; \phi_\calE\Big)$ 
        \State $\hat{\bsA}(t) = \texttt{sig} \Big( \bsH(t)\bsH(t)^\top \Big)$
        \For{$k \in [K]$}
            \State $\Tilde{\bsZ}_\texttt{cxt}^{(k)}(t) \gets \Tilde{\bsZ}_\texttt{cxt}(t) + \bsm \cdot \Big(\mathds{1}_{\bsX} -\bsM_{\bsX}^{(k)}(t) \Big)$
            \State $\hat{\bsZ}_{\texttt{tar}}^{(k)}(t) \gets g_{\texttt{node}} \Big(\Tilde{\bsZ}_\texttt{cxt}^{(k)}(t) ; \phi_\calV \Big)$ 
            \State $\calL_\texttt{node-spat}^{(t)} \gets \frac{1}{K} \sum_{k\in [K]} \calL_\texttt{MSE} \Big(\Hat{\bsZ}_{\texttt{tar}}^{(k)}(t), \bsZ_{\texttt{tar}}^{(k)}(t)\Big)$ \Comment{Spatial node latent reconstruction}
            \State $\calL_\texttt{adj-spat}^{(t)} \gets \frac{1}{K} \sum_{k\in [K]} \calL_\texttt{BCE} \Big(\Hat{\bsA}_{\texttt{tar}}^{(k)}(t), \bsA_{\texttt{tar}}^{(k)}(t)\Big)$ \Comment{Spatial adjacency matrix reconstruction}
        \EndFor

    \State $\calL_{\texttt{spat}}^{(t)} = \lambda_{\texttt{node}} \calL_{\texttt{node-spat}}^{(t)}+\lambda_{\texttt{adj}} \calL_{\texttt{adj-spat}}^{(t)} $

    \State $\calL_{\texttt{spat}} = \calL_{\texttt{spat}} + \calL_{\texttt{spat}}^{(t)}$ \Comment{Aggregate spatial reconstruction loss.}

    \State \texttt{/* Temporal reconstruction.*/}
        \State Sample $t_a \overset{\text{i.i.d.}}{\sim} U_{\text{int}}([0,t-\frac{1}{2}(\frac{\Gamma}{S}+1)]),\quad t_b \overset{\text{i.i.d.}}{\sim}U_{\text{int}}([t+\frac{1}{2}(\frac{\Gamma}{S}+1),T_\calG])$.
        
        \State $\Tilde{\bsZ}_{a,b}(t) \gets \bsW_{\texttt{T}} \Big[ \Tilde{\bsZ}(t_a)\,\Vert\,\Tilde{\bsZ}(t_b) \Big]$
        \State $\bar{\bsA}_{\texttt{tar}}(t) = \frac{1}{2}\Big(\texttt{sig}(\bsH(t_a)\bsH(t_b)^\top)+\texttt{sig}(\bsH(t_b)\bsH(t_a)^\top)\Big)$
        \For{$k \in [K]$}
            \State $\bar{\bsZ}_\texttt{cxt}^{(k)}(t) \gets \Tilde{\bsZ}_\texttt{cxt}(t) + \Tilde{\bsZ}_{a,b}(t) \odot  \Big( \mathds{1}_{\bsX} -\bsM_{\bsX}^{(k)}(t) \Big)$
            \State $\calL_{\texttt{node-temp}}^{(t)} \gets \frac{1}{K} \sum_{k\in [K]} \calL_{\texttt{MSE}} \Big( \bar{\bsZ}_{\texttt{tar}}^{(k)}(t), \bsZ_{\texttt{tar}}^{(k)}(t) \Big)$ \Comment{Temporal node latent reconstruction}
            \State $\calL_{\texttt{adj-temp}}^{(t)} = \frac{1}{K} \sum_{k\in [K]} \calL_{\texttt{BCE}} \Big(\bar{\bsA}_\texttt{tar}^{(k)}(t), \bsA_\texttt{tar}^{(k)}(t) \Big)$ \Comment{Temporal adjacency matrix reconstruction}
        \EndFor
    
    \State $\calL_\texttt{temp}^{(t)} \gets \lambda_{\texttt{node}} \calL_{\texttt{node-temp}}^{(t)} + \lambda_{\texttt{adj}} \calL_{\texttt{adj-temp}}^{(t)}$ 
     
    \State $\calL_\texttt{temp} \gets \calL_\texttt{temp} + \calL_\texttt{temp}^{(t)} $\Comment{Aggregate temporal reconstruction loss.}
\EndFor

\State  $\calL_\texttt{ST-JEMA} = \gamma\calL_\texttt{spat} + (1-\gamma)\calL_\texttt{temp},$
\State Update ${\theta}$, $\phi_{\calV}$ and $\phi_{\calE}$ by taking the gradient descent step with $\calL_\texttt{ST-JEMA}$ 
\State Update $\bar{\theta}$ using \gls{ema} based on $\theta$.
    
\EndFor 
\end{algorithmic}
\end{algorithm*}

\section{Constructing Dynamic Graphs from fMRI data}
\label{app:sec:constructdg}

\begin{figure}[t]
    \centering
    \includegraphics[width=0.48\textwidth]
    {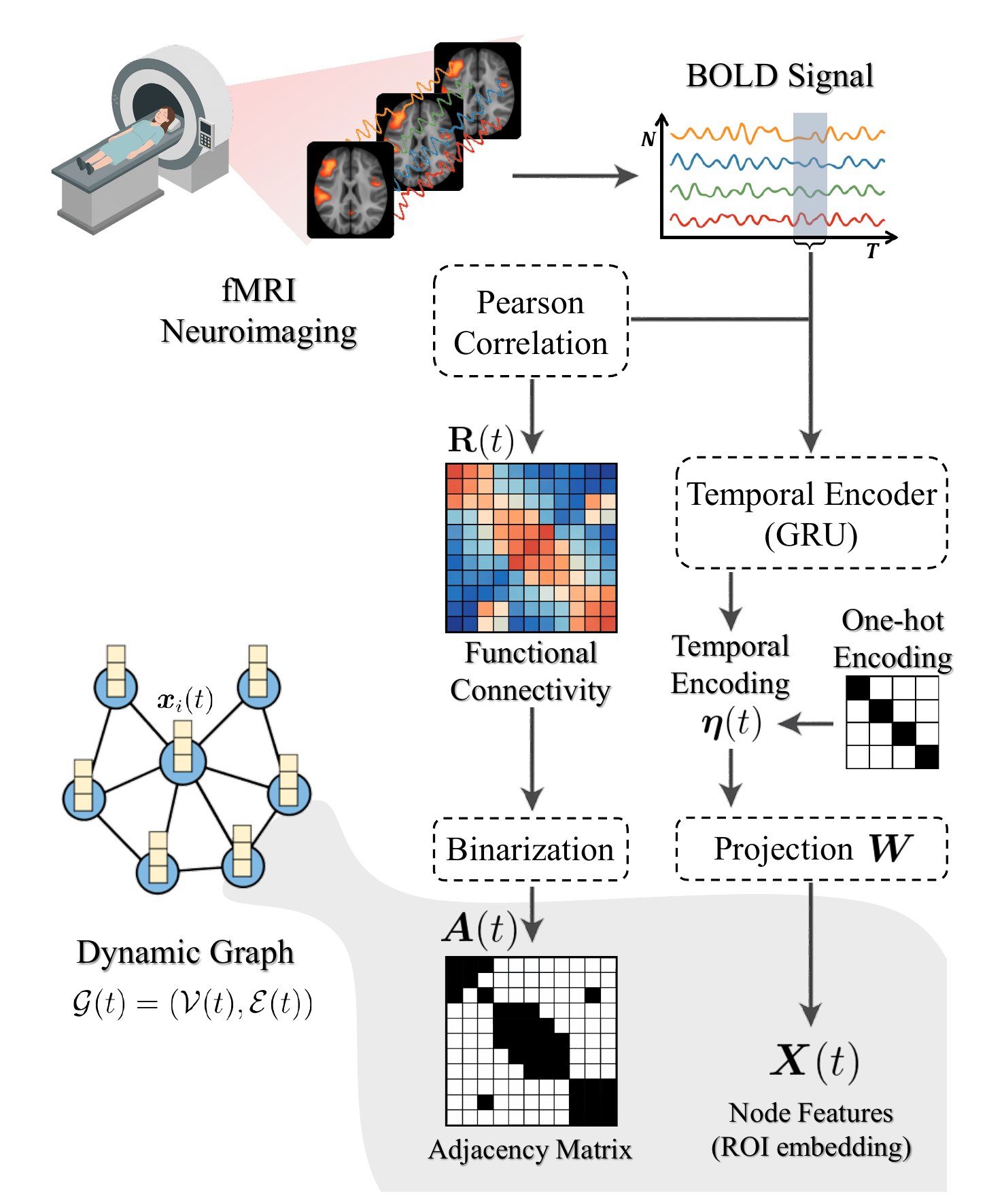}
    \caption{Overview of dynamic graph construction from \gls{fmri} data.}
    \label{fig:constructdg}
\end{figure}

\gls{fmri} data are complex and multidimensional, consisting of a sequence of brain images represented in 3D voxels collected over time. Each voxel within these images stands for a tiny volume in the brain, and the value of each voxel reflects the intensity of the \gls{bold} signal at that particular location and time.
In this study, we focus on neural activity within predefined \gls{roi}, Schaefer atlas~\citep{schaefer2018local}, leveraging its comprehensive partitioning of the cerebral cortex into functionally homogeneous regions. 
We construct the \gls{roi}-based \gls{bold} signal data matrix $\bsP \in \bbR^{N \times T_{max}}$, with $N$ representing the number of \gls{roi} and $T_{max}$ the time points, by averaging \gls{bold} signals within each \gls{roi}. This simplification retains essential details for subsequent dynamic functional connectivity analysis.

To analyze the \gls{bold} signal with \gls{gnn}, we construct dynamic graphs from the \gls{roi}-time-series matrix $\bsP$ of resting state \gls{fmri} (rs-fMRI) by defining node features $\bsX$ and the adjacency matrix $\bsA$. 
In this study, we utilize rs-\gls{fmri} because it allows for the exploration of spontaneous brain activity in the absence of task performance, offering insights into the brain's intrinsic functional connectivity from understanding fundamental brain networks to predicting phenotypes such as gender or disease presence.
For node feature $\bsX$, the \gls{roi}-time-series matrix $\bsP$ serves as the basis for generating node features that capture their temporal dynamics. Each node features $\bsx$ is combined spatial embeddings with learned temporal encodings using concatenation operation $\Vert$. Specifically, the feature vector of $i$-th node at time $t$ is formulated as:
\begin{equation}
\bsx_{i}(t) = \bsW[\bse_{i} \,\Vert \,\boldeta(t)],
\end{equation}
where $\bsW \in \bbR^{{d_v} \times (N + {d_\eta})}$ is a learnable parameter matrix, $\bse_i$ denotes the one-hot spatial embedding of the $i$-th node, and $\boldeta(t) \in \bbR^{d_\eta}$ is the temporal encoding obtained through a \gls{gru}~\citep{chung2014empirical}. Here, $d_\eta$ refers to the dimensionality of the temporal encoding.
This temporal encoding is designed to capture the dynamic changes in \gls{bold} signals within each \gls{roi} over time, thus incorporating temporal variations into the node features. Now, the final node feature matrix $\bsX(t)$ can be defined as a stack of node feature vectors at time $t$:
\begin{align}
    \bsX(t) = 
    \begin{bmatrix}
    \vert & & \vert \\
    \bsx_{1}(t) & \cdots & \bsx_{N}(t) \\
    \vert & & \vert 
    \end{bmatrix}. 
\end{align}
For the adjacency matrix $\bsA(t)$, 
we derive the \gls{fc} between \gls{roi} based on the temporal correlation of \gls{bold} signal changes, utilizing the \gls{roi}-timeseries matrix $\bsP$.
The \gls{fc} matrix at time $t$, $\bsR(t)$, is calculated as follows:
\begin{equation}
R_{ij}(t) = \frac{\text{Cov}(\bar{\bsp}_i(t), \bar{\bsp}_j(t))}
{\sqrt{\smash[b]{\text{Var}(\bar{\bsp}_i(t))}}\sqrt{\smash[b]{\text{Var}(\bar{\bsp}_j(t))}}} \in \bbR^{N \times N},
\end{equation}
where $\bar{\bsp}_i(t)$ and $\bar{\bsp}_j(t)$ are the \gls{bold} signal vectors for nodes $i$ and $j$ within the temporal windowed matrix $\bar{\bsP}(t) \in \bbR^{N \times \Gamma}$ from $\bsP$, characterized by a temporal window length of $\Gamma$ and stride $S$. Here, $t\in[T_\calG]$ where $T_\calG = \lfloor (T_\text{max} - \Gamma) / S \rfloor$. $\text{Cov}$ denotes the covariance, and $\text{Var}$ represents the variance operation. The adjacency matrix $\bsA(t) \in \{0, 1\}^{N \times N}$ is then obtained by applying a threshold to the correlation matrix $\bsR(t)$, wherein the top 30-percentile values are considered as significant connections (1), and all others are considered as non-significant (0). 
\section{Datasets}
\label{app:sec:dataset}

\begin{table*}[t]
    \caption{Phenotype information summary of the non-clinical and clinical rs-fMRI datasets}
    \label{tab:dataset}
    \begin{center}
    \begin{adjustbox}{width=\linewidth, totalheight=\textheight, keepaspectratio}
    \begin{tabular}{l  ccccc c cccc}
        \toprule       
        \multirow{2}{*}{Dataset} & \multicolumn{5}{c}{Non-clinical} && \multicolumn{4}{c}{Clinical} \\
        \cline{2-6}\cline{8-11}
        & UKB & ABCD & HCP-YA & HCP-D & HCP-A && HCP-EP & ABIDE & ADHD200 & COBRE \\
        % \midrule
        \cline{1-11}
        No. Subjects & 40\,913 & 9111 & 1093 & 632 & 723 && 176 & 884 & 668 & 173  \\
        Gender (F/M) & 21\,682 / 19\,231 & 4370 / 4741 & 594 / 499 & 339 / 293 & 405 / 318 && 67 / 109 & 138 / 746 & 242 / 426 & 43 / 130 \\
        Age (Min-Max) & 40.0-70.0 & 8.9-11.0 & 22.0-37.0 & 8.1-21.9 & 36.0-89.8 && 16.7-35.7 & 6.47-64.0 & 7.1-21.8 & 18.0-66.0 \\ 
        \multirow{2}{*}{Diagnosis (P/C)} &\multirow{2}{*}{-}&\multirow{2}{*}{-}&\multirow{2}{*}{-}&\multirow{2}{*}{-}&\multirow{2}{*}{-}&& SPR & ASD & ADHD & BPD / SPR / SZA \\
        &&&&&&& 120 / 56 & 408 / 476 & 280 / 389 & (9 / 69 / 11) / 84 \\
        \bottomrule
        
    \end{tabular}
    \end{adjustbox}
    \end{center}
\end{table*}

\subsection{\textbf{Data Preprocessing}}
\label{app:subsec:data_preprocessing}

We preprocessed all fMRI data to be registered with the standard Montreal Neurological Institute (MNI) space, ensuring consistency and comparability across datasets. The region of interest (ROI)-time-series matrices were extracted using the Schaefer atlas~\citep{schaefer2018local}, which identifies 400 distinct ROIs within the brain for analyzing functional connectivity.

From the time-series \gls{bold} signal $\bsP$ in each fMRI dataset, we construct two types of graphs: a static graph and a dynamic graph. While the dynamic graph is constructed as described in \cref{app:sec:constructdg}, for the static graph, we build the adjacency matrix $\bsA$ to represent the functional connectivity of \gls{roi} by windowing across all time stamps $\tau\in[T_{max}]$. Additionally, the node feature $\bsX$ is constructed using one-hot positional encoding, followed by a projection through the matrix $\bsW $.

\subsection{\textbf{Data Description}}

    \par{\textbf{UK Biobank (UKB)}~\citep{sudlow2015uk}}: 
    We utilized 40\,913 samples of resting-state fMRI from the UK Biobank, a comprehensive biomedical database of genetic, lifestyle, and health information from half a million participants. This dataset supports advanced medical research by offering insights into serious illnesses through diverse data collections, including genome sequencing, health record linkages, and functional MRI imaging.
    
    \par{\textbf{Adolescent Brain Cognitive Development (ABCD)}~\citep{casey2018adolescent}}: 
    A longitudinal study tracking nearly 12\,000 youth across 21 research sites to investigate the impact of childhood experiences on brain development and various behavioral, social, and health outcomes. We constructed 9,111 samples aged 9 to 11 years.
    
    \par{\textbf{Human Connectome Project Young Adults (HCP-YA)}~\citep{wu20171200}}: 
    Mapping the connectome of healthy individuals, this project provides a comprehensive database from 1200 young adults, incorporating advanced 3T and 7T MRI scanners for an exploration of brain connectivity and function. In our study, we specifically utilized rs-fMRI data from 1,093 subjects, following the pre-processing methodology described in \citet{kim2021learning}.
    \par{\textbf{Human Connectome Project Aging (HCP-A)}~\citep{bookheimer2019lifespan}}: 
    Investigating the effects of aging on brain structure and function, this segment of the Human Connectome Project includes 1200 adults aged 36 to 100+ years. It aims to uncover how brain connections change with age in healthy individuals using a 3T MRI scanner. For our analysis, we utilized rs-fMRI data from 724 samples aged 36 to 89 years.
    \par{\textbf{Human Connectome Project Development (HCP-D)}~\citep{somerville2018lifespan}}: 
    This dataset includes over 1300 participants ranging from 5 to 21 years old, exploring the dynamics of brain development from childhood through adolescence using a 3T MRI scanner. Specifically, our study utilizes rs-fMRI data from 632 individuals aged between 8 and 21 years.
    \par{\textbf{Human Connectome Project for Early Psychosis (HCP-EP)}~\citep{lewandowski2020neuroprogression}}:
    Focusing on early psychosis within the first three years of symptom onset, this dataset includes comprehensive imaging using a 3T MRI scanner, behavioral, clinical, cognitive, and genetic data from 400 subjects aged 16 to 35 years, including labels for schizophrenia (SPR). The objective is to identify disruptions in neural connections that underlie both brain function and dysfunction during a critical period with fewer confounds such as prolonged medication exposure. Our study employs rs-fMRI data from 176 subjects.
    \par{\textbf{Autism Brain Imaging Data Exchange (ABIDE)}~\citep{craddock2013neuro}}: 
    This dataset gathers rs-fMRI, anatomical, and phenotypic data from 1112 individuals across 17 international sites, aiming to enhance the understanding of Autism Spectrum Disorder (ASD) through collaborative research. For our study, we specifically utilized rs-fMRI data from 884 individuals, ranging in age from 6 to 64 years.
    \par{\textbf{ADHD200}~\citep{brown2012adhd}}: 
    This dataset initiative advances Attention Deficit Hyperactivity Disorder (ADHD) research by offering unrestricted access to 776 resting-state fMRI and anatomical datasets of individuals aged from 7 to 21 years, which includes 285 subjects with ADHD and 491 typically developing controls, from 8 international imaging sites. In our study, we utilized rs-fMRI data from 668 subjects.
    \par{\textbf{The Center for Biomedical Research Excellence (COBRE))~\citep{mayer2013functional}}
    }:
    We constructed rs-fMRI datasets comprising 173 samples, including 84 patients with schizophrenia (SPR), 9 with bipolar disorder (BPD), 11 with schizoaffective disorder (SZA), and 75 healthy controls, ranging in age from 18 to 66 years. This dataset aims to facilitate the exploration of neural mechanisms underlying various mental illnesses.
\section{Competing Methods}
\label{app:sec:baseline}

To compare with our proposed \gls{ours}, we evaluated eight competing algorithms, each designed for pre-training \gls{gnn} encoders using \gls{ssl} methods, excluding Random-Init:
    \par{ \textbf{Random-Init}}: Training from scratch on the downstream dataset without the pre-training step using the upstream dataset. We present two sets of results for Random-Init, specifically when using a static graph as input and a dynamic graph as input, as elaborated in \cref{app:subsec:data_preprocessing}.
    \par{ \textbf{DGI}~\citep{velivckovic2018deep}}: A contrastive \gls{ssl} method for static graph data that learns mutual information between node and graph representations.
    \par{ \textbf{SimGRACE}~\citep{xia2022simgrace}}: A contrastive \gls{ssl} method for static graph data that utilizes model parameter perturbation to generate different views instead of using graph augmentations.
    \par{ \textbf{GAE}~\citep{kipf2016variational}}: A generative \gls{ssl} method for static graph data that utilizes an autoencoder framework. GAE reconstructs a graph's original adjacency matrix by leveraging node representations.
    \par{ \textbf{VGAE}~\citep{kipf2016variational}}: A generative \gls{ssl} method for static graph data that improves GAE by using a variational autoencoder in the encoder. 
    \par{ \textbf{GraphMAE}~\citep{hou2022graphmae}}: A generative \gls{ssl} method for static graph data that focuses on node reconstruction with masked autoencoder loss.
    \par{ \textbf{ST-DGI}~\citep{opolka2019spatio}}: A contrastive \gls{ssl} method for dynamic graph data that enhances DGI by optimizing the mutual information between node embeddings $\bsZ(t)$ and features at future k-time steps $\bsZ(t+k)$.
    \par{ \textbf{ST-MAE}~\citep{choi2023generative}}: A generative \gls{ssl} method specifically designed for \gls{fmri} data. ST-MAE focuses on dynamic \gls{fc} by utilizing a masked autoencoder  framework that accounts for temporal dynamics.
\section{Experimental Details}
\label{app:sec:details}

In this section, we detail the implementation of the models for \gls{ours} and training pipelines for upstream and downstream datasets.

\subsection{Architecture Details}

In this research, we utilize a 4-layer Graph Isomorphism Network (GIN)~\citep{xu2018powerful} as the \gls{gnn} encoder and the Squeeze-Excitation READOUT (\texttt{SERO}) readout module in the fine-tuning process following the approach introduced by \citet{kim2021learning}. To ensure that each decoder effectively reconstructs target components from context node representations, we employ MLP-Mixer~\citep{tolstikhin2021mlp} for both node and edge decoders. Here, we provide detailed architectural information about GIN, \texttt{SERO}, and MLP-Mixer.

\textbf{Graph Isomorphism Network (GIN)}~\citep{xu2018powerful}: 

Graph Isomorphism Networks, are specialized spatial \gls{gnn} optimized for graph classification, achieving performance comparable to the Weisfeiler-Lehman (WL) test~\citep{leman1968reduction}. The GIN architecture aggregates neighboring node features through summation and employs an MLP for feature transformation, enabling the learning of injective mappings. This unique structure enables GINs to adeptly capture complex graph patterns, proving particularly effective in analyzing \gls{fmri} data for tasks such as gender classification from rs-\gls{fmri} data~\citep{kim2020understanding}. The mathematical formulation of GIN at layer $l$ is as follows:
\begin{align}
  \bsz_i^{(l)} = \text{MLP}^{(l)}\left((1 + \epsilon^{(l)}) \cdot \bsz_i^{(l-1)} + \sum_{j \in \calN(i)} \bsz_j^{(l-1)}\right),    
\end{align}
where $\epsilon^{(k)}$ is a parameter optimized during training.
The application of GINs to \gls{fmri} data analysis marks a significant progression, offering refined insights into the spatial and temporal dynamics of brain activity.

\textbf{Squeeze-Excitation READOUT (SERO)}~\citep{kim2021learning}:
\texttt{SERO} is the noble readout module that adapts the concept of attention mechanisms from Squeeze-and-Excitation Networks~\citep{hu2018squeeze}, adjusting it to apply attention across the node dimension instead of the channel dimension for \glspl{gnn}. This modification enhances the focus on the structural and functional properties of nodes within brain graphs. The attention mechanism is mathematically defined as follows: 
\begin{align}
\bsa_{\text{space}}(t) = \texttt{sig} \big( \bsW_2 \sigma(\bsW_1 \bsZ(t) \Phi_{\text{mean}}) \big),
\end{align}
where $\Phi_{\text{mean}} = [1/N, ..., 1/N]$ is a mean vector for mean pooling, 
$\bsW_1 \in \bbR^{d_{\text{enc}} \times d_{\text{enc}}}, \bsW_2 \in \bbR^{N \times d_{\text{enc}}}$ are learnable parameter matrices, and $\sigma$ is the nonlinear activation function such as ReLU.  Finally, The graph representation at time $t$ is derived by applying attention weights $\bsa_{\text{space}}$(t) to each node of $Z(t)$, effectively weighting the importance of each node's features in the overall graph representation as follows:
\begin{align}
    \bsz_{G}(t) = \bsZ(t) \bsa_{\text{space}}(t).
\end{align}
This readout $\bsz_{G}(t)$ is used when the model is fine-tuned on the downstream task. 
When integrated into the STAGIN~\citep{kim2021learning} framework, this module significantly enhances performance in gender and task classification on \gls{fmri} data, demonstrating advancements over existing models.

\textbf{MLP-Mixer}~\citep{tolstikhin2021mlp}:
MLP-Mixer adopts a purely MLP-based architecture, diverging from conventional convolution and self-attention mechanisms, comprising of token-mixing and channel-mixing \glspl{mlp}.
In this setup, we utilize the token-mixing \glspl{mlp} to encourage interactions among nodes, while the channel-mixing \glspl{mlp} facilitates feature communication across the feature dimension of node representations.
For a graph with node representations $\mathbf{Z} \in \mathbb{R}^{N \times D}$, where $N$ is the number of nodes and $D$ is the dimension of node features, the MLP-Mixer operates as follows:
\begin{align}
\begin{gathered}
    \bsY_{*,i} = \bsZ_{*,i} + \bsW_{2} \texttt{GELU}(\bsW_{1} \texttt{LN}(\bsZ)_{*,i}), \, i = \{1,...,N\},\\
    \hat{\bsZ}_{j,*} = \bsY_{j,*} + \bsW_{4} \texttt{GELU}(\bsW_{3} \texttt{LN}(\bsY)_{j,*}), \, j = \{1,...,D\},
\end{gathered}
\end{align}
where \texttt{GELU}~\citep{hendrycks2016gaussian} is a non-linear activation function and \texttt{LN}~\citep{ba2016layer} is a layer normalization. 
And $\bsW_{1} \in \bbR^{\bar{N} \times N}$, $\bsW_{2} \in \bbR^{N \times \bar{N}}$, $\bsW_{3} \in \bbR^{\bar{D} \times D}$, and $\bsW_{4} \in \bbR^{D \times \bar{D}}$ are learnable parameters for each \glspl{mlp} with hidden dimension of $\bar{N}$ and $\bar{D}$.
This adaptation allows the MLP-Mixer to ensure communication between nodes for the reconstruction of targeted node representations from the aggregated context of the graph. 

\subsection{Experimental Details}

\subsubsection*{Dynamic Graph Construction Settings}
For the upstream dataset in our \gls{ssl} methods, we utilize the UKB dataset with a window size $\Gamma = 50$ and a window stride $S = 16$ for constructing dynamic graphs. We split the upstream dataset into train (80\%) / validation (10\%) / test (10\%) partitions and selected the pre-trained model for each \gls{ssl} method based on the best-performing upstream \gls{ssl} task on the test dataset. For downstream tasks, we construct dynamic graphs from \gls{fmri} datasets such as ABCD, HCP-YA, HCP-A, HCP-D, and HCP-EP are processed with a window size of 50 and a stride of 16. For the ABIDE, ADHD200, and COBRE datasets, we use a window size of 16 and a stride of 3 for consistency of \gls{fmri} scanning time. 
Following the approach by \citet{kim2021learning}, we also randomly sliced the time dimension of the \gls{roi}-time-series \gls{bold} signal matrix $\bsP$, setting dynamic lengths of 400 for UKB, HCP-A, HCP-D, and HCP-EP; 380 for ABCD; 600 for HCP-YA; 70 for ABIDE and ADHD200; and 120 for COBRE.

\subsubsection*{Training Configuration and Optimization}
We optimize the \gls{gnn} encoder in two stages: (1) self-supervised pretraining on the upstream dataset, and (2) supervised finetuning on downstream tasks.

Pre-training employed a cosine decay learning schedule over 10\,000 steps with a batch size of 16. Fine-tuning optimized the encoder with the task-specific head, using orthogonal regularization and a one-cycle scheduler for 30 epochs with a batch size of 32. Dynamic graphs integrated a mean pooling module for final representation processing. Classification tasks used cross-entropy loss, while age regression utilized \gls{mse} loss with normalized age data. The training utilized AMD EPYC processors, with NVIDIA RTX A6000 for pre-training and GeForce RTX 3090 for fine-tuning and ablation studies.

Optimization hyperparameters include a learning rate of $1 \times 10^{-4}$ for ST-DGI, $5 \times 10^{-4}$ for ST-MAE, and $5 \times 10^{-5}$ for other \gls{ssl} methods, with a consistent weight decay of $1 \times 10^{-4}$ in the pre-training phase.
For finetuning, a learning rate of $5 \times 10^{-4}$ is used for gender and psychiatric diagnosis classification tasks, and $1 \times 10^{-3}$ for the age regression task, selecting the best-performing model within weight decays of $\{1 \times 10^{-3}, 5 \times 10^{-4}, 1 \times 10^{-4}\}$. For Random-Init, a learning rate of $1 \times 10^{-3}$ and weight decay of $1 \times 10^{-4}$ are used across all tasks.

Additional hyperparameters for each \gls{ssl} method are summarized in ~\cref{tab:supp_hyperparams}.
\begin{table}[ht]
\centering
\caption{Additional hyperparameters used for each SSL method.}
\begin{tabular}{ll}
\toprule
\textbf{Method} & \textbf{Key Hyperparameters} \\
\midrule
SimGRACE & Perturbation scale $\eta = 1.0$ \\
GraphMAE & Masking ratio $\alpha = \{30, 50, 70\}$ (\%) \\
ST-DGI & Temporal steps $k = \{1, 3, 6\}$ \\
ST-MAE & Masking ratio $\alpha = \{30, 50, 70\}$ (\%) \\
\textbf{ST-JEMA} & Block mask $\alpha_{\min\text{--}\max} = 10\text{--}30$ (\%), $\gamma=0.5$, \\
& $\lambda_{\text{node}} = 1.0$, $\lambda_{\text{edge}} = \{10^{-3}, 10^{-4}, 10^{-5}, 10^{-6}\}$ \\
\bottomrule
\end{tabular}
\label{tab:supp_hyperparams}
\end{table}

\section{Additional Experiments}
\label{app:additional_experiments}

\begin{table}[t]
\centering
\caption{Linear probing performance of psychiatric diagnosis classification. The performance is evaluated using the \gls{auroc} score, with higher scores indicating better performance.}
\label{app:tab:linear_diagnosis}
\begin{tabular}{l  c c c c }
\toprule
\multirow{2}{*}{Pre-training} &  \multicolumn{4}{c}{AUROC ($\uparrow$) } \\
\cline{2-5}
& HCP-EP & ABIDE & ADHD200 &  COBRE \\
\cline{1-5}
ST-DGI & 70.66 & 53.87 & 53.32 & 63.73 \\
ST-MAE & 70.78 & 55.09 & 52.66 & 63.97 \\
\rowcolor{skyblue}
ST-JEMA & $\mathbf{72.97}$ & $\mathbf{58.30}$ & $\mathbf{54.44}$ & $\mathbf{66.39}$ \\
\bottomrule
\end{tabular}
\end{table}
\begin{table}[t]
\centering
\caption{Multi-task learning performance on ABIDE and ADHD200 datasets. Here, (-) represents a degree of performance drop, and (+) indicates the degree of performance improvement compared to the performance in each single-task learning scenario.}
\label{app:tab:multitask}

\begin{tabular}{l  ccc}
\toprule
\multirow{3}{*}{Pre-training}
& \multicolumn{3}{c}{ABIDE}\\ \cline{2-4}
& Gender & Age & Diagnosis \\ 
\cline{2-4}
& \gls{auroc} ($\uparrow$) & \gls{mae} ($\downarrow$) &\gls{auroc} ($\uparrow$) \\ 
\midrule
ST-DGI & 65.00 ({-9.87}) & 4.24 ({-0.08})& 59.50 ({-9.36})\\
ST-MAE & 62.86 ({-11.88})& 4.11 ({-0.01})& 58.25 ({-11.48})\\
\rowcolor{skyblue}
ST-JEMA & $\mathbf{65.60}$ ({-9.78})& $\mathbf{4.06}$ ({+0.02})& $\mathbf{66.10}$ ({-5.39})\\  
\midrule
\multirow{3}{*}{Pre-training}
& \multicolumn{3}{c}{ADHD200}\\ \cline{2-4}
& Gender & Age & Diagnosis \\ 
\cline{2-4}
& \gls{auroc} ($\uparrow$) & \gls{mae} ($\downarrow$) &\gls{auroc} ($\uparrow$) \\ 
\midrule
ST-DGI & 60.06 ({-7.98})& 1.99 ({-0.08})& 55.61 ({-0.85})\\
ST-MAE &  60.98 ({-10.03})& $\mathbf{1.89}$ ({-0.03})& 50.45 ({-6.32})\\
\rowcolor{skyblue}
ST-JEMA & $\mathbf{68.29}$ ({-3.65})& $1.91$ ({-0.05})& $\mathbf{59.84}$ ({+0.95})\\  

\bottomrule
\end{tabular}
\end{table}

\subsection{Linear Probing Performance}
\label{app:subsec:linearprob}

A technique that can indirectly reflect whether the representations acquired through \gls{ssl} in the pre-training phase contain valuable information for the downstream task is to perform linear probing on the downstream task and measure its performance~\citep{alain2016understanding}. In this context, linear probing refers to the learning process wherein the pre-trained encoder module in the model is frozen, and training is conducted solely on the task-specific head $h(\cdot,\bomega)$ using the downstream dataset.

According to the results in \cref{app:tab:linear_diagnosis}, \gls{ours} exhibited higher AUROC performance than ST-MAE, suggesting that \gls{ours} captures more semantically rich information beneficial for psychiatric diagnosis classification. On the contrary, ST-DGI, a \gls{ssl} method based on contrastive learning, exhibited lower performance than generative approaches. This suggests that the representations learned by ST-DGI might not be as beneficial for psychiatric diagnosis classification tasks compared to generative-based \gls{ssl} methods.

\subsection{Multi-task Learning Performance}
\label{app:subsec:multitask}

To further verify that \gls{ours} learns more informative representations compared to other baseline models, we conduct a multi-task learning experiment. In this setting, we concurrently train on three distinct tasks $-$ gender classification, age regression, and diagnosis classification $-$ within each dataset. To facilitate the simultaneous training of these three tasks, we employ three separate task-specific heads and fine-tune the pre-trained encoder and task-specific heads concurrently, utilizing aggregated loss from the three tasks.

\cref{app:tab:multitask} clearly shows that \gls{ours} achieved the best performance across all tasks compared to other baselines except for the age regression task in the ADHD200 dataset. Notably, \gls{ours} exhibits the smallest overall performance decline for each task when compared to the performance in the single-task learning scenario, outperforming other baseline models. In particular, regarding the age regression task in the ABIDE dataset and diagnosis classification task in ADHD200, \gls{ours} demonstrates an improvement in performance in the multi-task learning scenario compared to its performance in the single-task learning scenario. These findings highlight that the pre-trained encoder in \gls{ours} has effectively captured mutual information beneficial for addressing diverse downstream tasks.

\subsection{Robustness on Missing Data Scenarios}
\label{app:sub:ablation_missingdata}

\begin{figure}[t]
    
        \centering
        \includegraphics[width=0.5\linewidth]{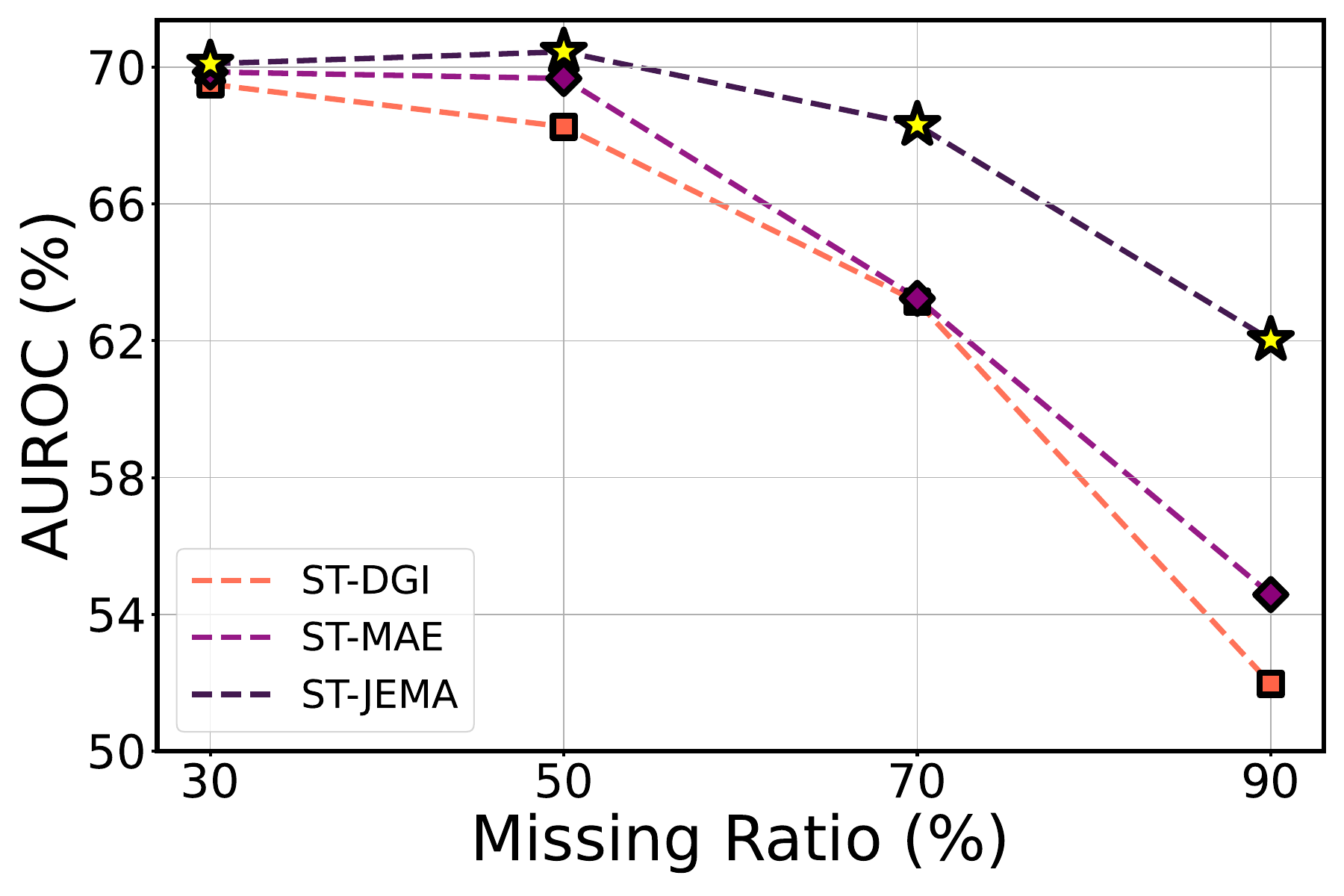}
        \caption{
        Ablation results on temporal missing data scenarios on psychiatric diagnosis classification task on ABIDE dataset. 
        We designed temporal missing data scenarios by randomly masking the \gls{bold} signal from \gls{fmri} data along the time axis, adjusting the missing ratio from 30\% to 90\%, and then measured the \gls{auroc} score to evaluate the effectiveness of \gls{ssl} in capturing the temporal dynamics.}
        \label{fig:missing_data}
        
\end{figure}

To further confirm that \gls{ours} successfully captured the temporal dynamics in dynamic \gls{fc} of \gls{fmri} data, we perform ablation experiments on the scenario of temporal missing data for the downstream dataset. Here, the temporal missing data scenario refers to assuming that a certain portion of timesteps is missing from the original data for each sample. To create the missing part from the original downstream training data, we mask randomly selected timesteps in \gls{bold} signal according to a predefined missing ratio. We conduct experiments on the psychiatric diagnosis classification task using the ABIDE dataset, adjusting the missing ratio from 30\% to 90\%. 
As demonstrated in \cref{fig:missing_data}, we observed a decline in fine-tuning performance through all methods as the missing ratio increased. This indicates that the loss of temporal flow hinders the accurate prediction of psychiatric disorders, highlighting the importance of learning temporal dynamics for precise predictions. While ST-DGI and ST-MAE exhibited significant performance drops as the missing ratio increased, \gls{ours} maintained relatively better classification performance even with 90\% of the data missing along the time axis. This suggests that our proposed \gls{ssl} method enables the model to effectively learn temporal dynamics, facilitating the learning of pre-trained representations useful for downstream tasks such as psychiatric diagnosis classification.

\subsection{Impact of Pre-Training Data Size}
\label{app:sub:ssl_data_size}

\begin{figure}[t]
        \centering
        \includegraphics[width=0.5\linewidth]{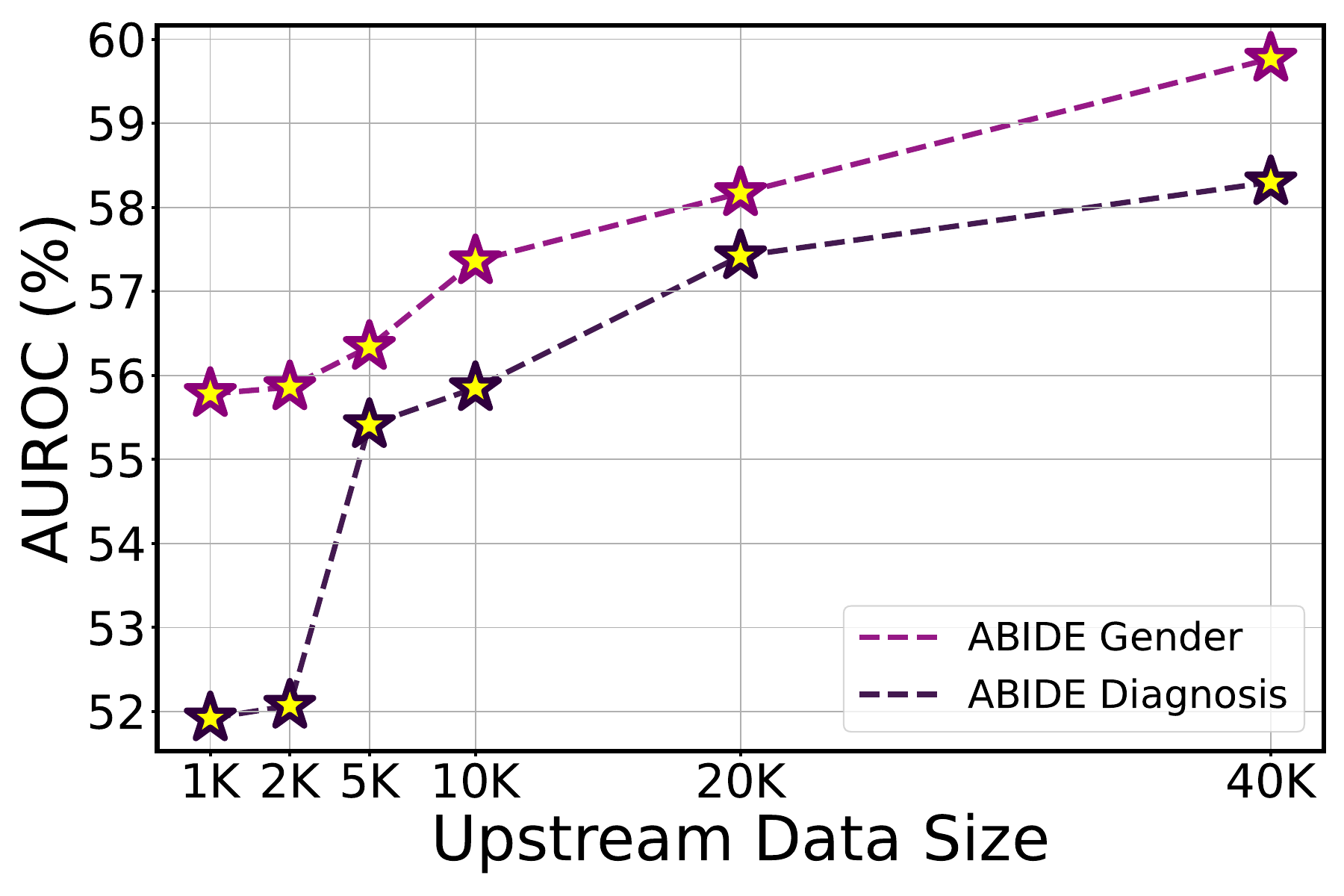}
        \caption{Ablation study on the impact of sample size for \gls{ssl} on the ABIDE dataset. We evaluated learned representations for downstream tasks by training a linear probing model across varying numbers of samples.}
        \label{fig:ssl_data_size}
\end{figure}

To confirm the impact of the quantity of upstream UKB data on the quality of the pre-trained representation by the \gls{gnn} encoder, we perform an ablation experiment by varying the size of the pre-training data. To systematically evaluate the impact of varying training data sizes on the effectiveness of pre-trained representations in solving downstream tasks, we conducted linear probing on the gender classification task and the diagnosis classification task using the ABIDE dataset. Notably, linear probing preserves the pre-trained representations even after the fine-tuning process which effectively shows the quality of pre-trained representations.
\cref{fig:ssl_data_size} clearly indicates that the performance improves as the size of the training data increases. This result demonstrates that \gls{ours} successfully learns more meaningful representations as the training dataset size increases. It suggests that \gls{ours} is adept at acquiring high-quality representations from the extensive UKB upstream dataset.

\subsection{Ablation on Block Mask Ratio}
\label{app:subsub:ablation_block_mask}

\begin{figure}[t]
    \centering
    
        \includegraphics[width=0.5\linewidth]{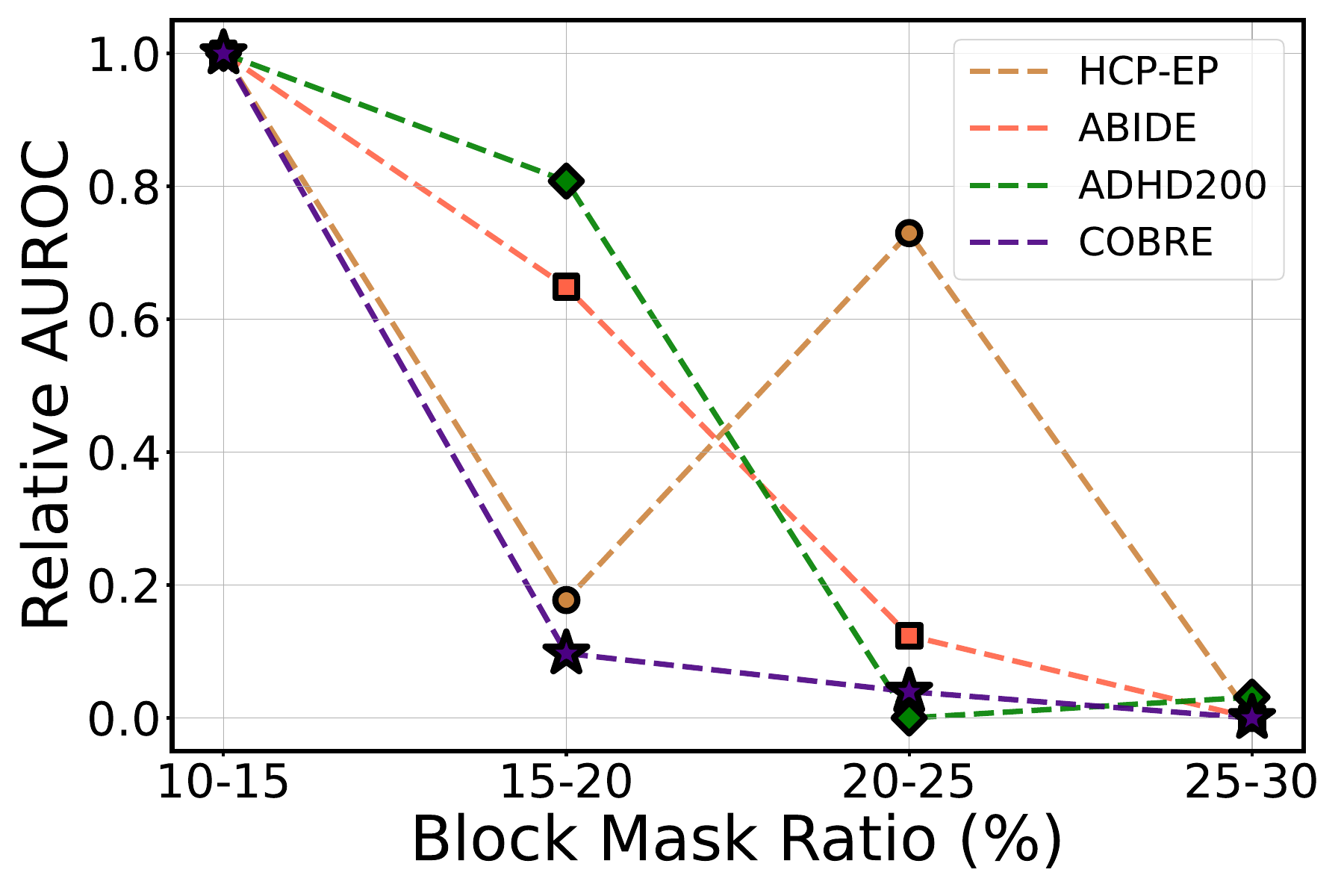}
        \caption{Ablation study on the block mask ratio of \gls{ours} for psychiatric diagnosis classification in clinical data. The x-axis, representing the block mask ratio, signifies the relative ratio of the minimum and maximum values within a specified range relative to the total size. The size of each block mask is randomly sampled from this range. To provide the outcomes across all datasets simultaneously, we present the relative AUROC results. Here, we calculate the relative AUROC by normalizing using the minimum and maximum AUROC results obtained from different block mask ratio ranges.}
        \label{fig:block_mask_ratio}
    
\end{figure}
In the masking process of target points in \gls{ours}, selecting the appropriate block mask ratio is crucial. It is essential to strike a balance that ensures sufficient information for reconstructing the target representation, while also making the problem challenging enough to prevent the \gls{gnn} encoder from learning trivial representations.
To analyze the appropriate block mask ratio in \gls{ours}, we perform an ablation experiment on the block mask ratio using the diagnosis classification task.

In \cref{fig:block_mask_ratio}, one can see that utilizing $10\%\sim15\%$ is the optimal block mask ratio across all datasets. The trends are consistent across all datasets except for HCP-EP, where the performance improves with a block mask ratio of $20\%\sim25\%$ compared to $15\%\sim20\%$. Since we employ $4$ masks for each timestep $t\in[T_{\calG}]$, the optimal block mask ratio might change if the number of masks per each node and adjacency matrix is altered.

\subsection{Decoder Architecture}
\label{app:subsub:ablation_decoder}

\begin{table}[t]
\centering
\caption{Ablation results on different architecture for decoder module with diagnosis classification task on ABIDE dataset. The performance is evaluated using the AUROC score, with higher scores indicating better performance.}
\label{app:tab:ablation_decoder}
\begin{tabular}{ l c}
\toprule
Decoder architecture &  AUROC ($\uparrow$) \\
\midrule
MLP & 64.76 \\
GIN~\citep{xu2018powerful} & 66.35 \\
\midrule
\rowcolor{skyblue}
MLP-Mixer~\citep{tolstikhin2021mlp} & $\mathbf{71.49}$ \\
\bottomrule
\end{tabular}
\end{table}

As described in \cref{main:sec:method}, We implemented the MLP-Mixer~\citep{tolstikhin2021mlp} to reconstruct the target node representation by integrating information from context node representations. Another approach to aggregate representations from different nodes is by using a \gls{gnn} decoder. GraphMAE employed a more expressive \gls{gnn} decoder instead of a simple MLP to prevent learning representations nearly identical to the low-level semantics of input features, thus aiming to capture higher-level semantics.

To determine which decoder structure is most effective for our methodology, we compared an MLP module, a one-layer GIN module, and a one-layer MLP-Mixer module. The MLP module does not utilize information from other node representations for reconstructing the target node and hence does not use a learnable token for node representation; instead, it reconstructs the target node representation using the corresponding node representation obtained from the \gls{gnn} encoder. As seen in \cref{app:tab:ablation_decoder}, the MLP decoder exhibited the lowest performance, possibly due to the pre-trained representations being of low-level semantics. While the GIN module performed better, its performance was similar to that of a random initialization, indicating limited benefit. In contrast, our method using the MLP-Mixer facilitated interactions among context node representations, thereby learning representations with higher-level semantics.

\section{Additional Related Works}
\label{app:sec:addtional_relatedwork}

\subsection{Contrastive Self-supervised Learning on Graph}

Contrastive methods focus on aligning representations derived from different `views' of input graphs, known as positive pairs, typically generated through data augmentations or multiple snapshots of the same graph. 
The primary goal is to maximize the agreement between these positive pairs of samples, thereby enhancing the model's ability to capture meaningful relationships within the data. To achieve this, a projection head is commonly employed to transform the encoder's representations into an optimized feature space, enhancing discriminative power and learning stability while mitigating overfitting and preventing representation mode collapse~\citep{jing2021understanding, gupta2022understanding}.
DGI~\citep{velivckovic2018deep} pioneered in graph contrastive \gls{ssl}, aiming to maximize the mutual information between local and global graph representations. Subsequently, GraphCL~\citep{you2020graph} introduced graph augmentations to create different views for contrastive learning. 
However, the effectiveness of these augmentations can vary across different domains and sometimes restrict the graph structure. In response to these challenges, alternative strategies have been developed to bypass the need for complex augmentations. 
SimGCL~\citep{yu2022graph} adopted a novel approach by adding noise to the embedding space to generate contrastive views, while SimGRACE~\citep{xia2022simgrace} simplified the process by leveraging correlated views from perturbed \gls{gnn} parameters. 

\subsection{Generative Self-supervised Learning on Graph}

Generative approaches aim to reconstruct the original input by leveraging the encoded representations. Employing a decoder, these methods are tasked with reconstructing the original graph structure from the encoder’s embeddings. By optimizing the \gls{gnn} model for precise reconstruction of the graph structure, generative methods enable the learning of representations that inherently capture the structural relationships in graph data. 
VGAE~\citep{kipf2016variational}, a pioneer work in generative approach, introduces strategies to reconstruct a graph's adjacency matrix from node embeddings. Utilizing variational autoencoders~\citep{kingma2013auto}, VGAE demonstrates better performance in link prediction than conventional unsupervised learning methods. DeepWalk~\citep{perozzi2014deepwalk}, highlighting their effectiveness in identifying important patterns in graph-structured data without labeled graph data. 
GraphMAE~\citep{hou2022graphmae} adopts a novel approach to \gls{ssl} in graph data by implementing a masked autoencoder for node feature reconstruction. GraphMAE demonstrates notable advantages over traditional contrastive learning techniques in tasks such as node and graph classification, showing evidence that generative self-supervised is also effective in the graph domain beyond the computer vision task.  

\end{document}